\documentclass[letterpaper]{article} 
\usepackage{aaai25}  
\usepackage{times}  
\usepackage{helvet}  
\usepackage{courier}  
\usepackage[hyphens]{url}  
\usepackage{graphicx} 
\usepackage{natbib}  
\usepackage{caption} 
\usepackage{algorithm}
\usepackage{algorithmic}
\usepackage{booktabs}
\usepackage{amsmath}
\usepackage{amssymb}
\usepackage{caption}
\usepackage{subcaption}
\usepackage{cuted}
\usepackage{microtype}

\usepackage{newfloat}
\usepackage{listings}
\DeclareCaptionStyle{ruled}{labelfont=normalfont,labelsep=colon,strut=off} 
\floatstyle{ruled}
\newfloat{listing}{tb}{lst}{}
\floatname{listing}{Listing}
\let\titleold\title
\renewcommand{\title}[1]{\titleold{#1}\newcommand{\thetitle}{#1}}
\def\maketitlesupplementary
   {
   \newpage
       \twocolumn[
        \centering
        \Large
        \textbf{\thetitle}\\
        \vspace{0.5em}Supplementary Material \\
        \vspace{1.0em}
       ] 
   }

\nocopyright 

\title{\model{}: Progressively Refined View Synthesis\\from 3D Lifting with Volume-Triplane Representations}

\author {
    Tung Do\textsuperscript{\rm 1},
    Thuan Hoang Nguyen\textsuperscript{\rm 1},
    Anh Tuan Tran\textsuperscript{\rm 1},
    Rang Nguyen\textsuperscript{\rm 1},
    Binh-Son Hua\textsuperscript{\rm 1,\rm 2}
}
\affiliations {
    \textsuperscript{\rm 1} VinAI Research, Vietnam \\ 
    \textsuperscript{\rm 2} Trinity College Dublin, Ireland\\
}

\usepackage{bibentry}
\usepackage{xcolor}
\begin{document}

\newcommand{\son}[1]{{\textcolor{red}{#1}}}
\newcommand{\tung}[1]{{\textcolor{orange}{#1}}}
\newcommand{\thuan}[1]{{\textcolor{cyan}{#1}}}
\newcommand{\rang}[1]{{\textcolor{blue}{#1}}}

\def\mA{\mathcal{A}}
\def\mB{\mathcal{B}}
\def\mC{\mathcal{C}}
\def\mD{\mathcal{D}}
\def\mE{\mathcal{E}}
\def\mF{\mathcal{F}}
\def\mG{\mathcal{G}}
\def\mH{\mathcal{H}}
\def\mI{\mathcal{I}}
\def\mJ{\mathcal{J}}
\def\mK{\mathcal{K}}
\def\mL{\mathcal{L}}
\def\mM{\mathcal{M}}
\def\mN{\mathcal{N}}
\def\mO{\mathcal{O}}
\def\mP{\mathcal{P}}
\def\mQ{\mathcal{Q}}
\def\mR{\mathcal{R}}
\def\mS{\mathcal{S}}
\def\mT{\mathcal{T}}
\def\mU{\mathcal{U}}
\def\mV{\mathcal{V}}
\def\mW{\mathcal{W}}
\def\mX{\mathcal{X}}
\def\mY{\mathcal{Y}}
\def\mZ{\mathcal{Z}}
\def\bbN{\mathbb{N}}
\def\bbR{\mathbb{R}}
\def\bbP{\mathbb{P}}
\def\bbQ{\mathbb{Q}}
\def\bbE{\mathbb{E}}
\def\1n{\mathbf{1}_n}
\def\0{\mathbf{0}}
\def\1{\mathbf{1}}
\def\A{{\bf A}}
\def\B{{\bf B}}
\def\C{{\bf C}}
\def\D{{\bf D}}
\def\E{{\bf E}}
\def\F{{\bf F}}
\def\G{{\bf G}}
\def\H{{\bf H}}
\def\I{{\bf I}}
\def\J{{\bf J}}
\def\K{{\bf K}}
\def\L{{\bf L}}
\def\M{{\bf M}}
\def\N{{\bf N}}
\def\O{{\bf O}}
\def\P{{\bf P}}
\def\Q{{\bf Q}}
\def\R{{\bf R}}
\def\S{{\bf S}}
\def\T{{\bf T}}
\def\U{{\bf U}}
\def\V{{\bf V}}
\def\W{{\bf W}}
\def\X{{\bf X}}
\def\Y{{\bf Y}}
\def\Z{{\bf Z}}
\def\a{{\bf a}}
\def\b{{\bf b}}
\def\c{{\bf c}}
\def\d{{\bf d}}
\def\e{{\bf e}}
\def\f{{\bf f}}
\def\g{{\bf g}}
\def\h{{\bf h}}
\def\i{{\bf i}}
\def\j{{\bf j}}
\def\k{{\bf k}}
\def\l{{\bf l}}
\def\m{{\bf m}}
\def\n{{\bf n}}
\def\o{{\bf o}}
\def\p{{\bf p}}
\def\q{{\bf q}}
\def\r{{\bf r}}
\def\s{{\bf s}}
\def\t{{\bf t}}
\def\u{{\bf u}}
\def\v{{\bf v}}
\def\w{{\bf w}}
\def\x{{\bf x}}
\def\y{{\bf y}}
\def\z{{\bf z}}
\def\balpha{\mbox{\boldmath{$\alpha$}}}
\def\bbeta{\mbox{\boldmath{$\beta$}}}
\def\bdelta{\mbox{\boldmath{$\delta$}}}
\def\bgamma{\mbox{\boldmath{$\gamma$}}}
\def\blambda{\mbox{\boldmath{$\lambda$}}}
\def\bsigma{\mbox{\boldmath{$\sigma$}}}
\def\btheta{\mbox{\boldmath{$\theta$}}}
\def\bomega{\mbox{\boldmath{$\omega$}}}
\def\bxi{\mbox{\boldmath{$\xi$}}}
\def\bnu{\mbox{\boldmath{$\nu$}}}
\def\bphi{\mbox{\boldmath{$\phi$}}}
\def\bmu{\mbox{\boldmath{$\mu$}}}
\def\bDelta{\mbox{\boldmath{$\Delta$}}}
\def\bOmega{\mbox{\boldmath{$\Omega$}}}
\def\bPhi{\mbox{\boldmath{$\Phi$}}}
\def\bLambda{\mbox{\boldmath{$\Lambda$}}}
\def\bSigma{\mbox{\boldmath{$\Sigma$}}}
\def\bGamma{\mbox{\boldmath{$\Gamma$}}}
\newcommand{\myprob}[1]{\mathop{\mathbb{P}}_{#1}}
\newcommand{\myexp}[1]{\mathop{\mathbb{E}}_{#1}}
\newcommand{\mydelta}[1]{1_{#1}}
\newcommand{\myminimum}[1]{\mathop{\textrm{minimum}}_{#1}}
\newcommand{\mymaximum}[1]{\mathop{\textrm{maximum}}_{#1}}
\newcommand{\mymin}[1]{\mathop{\textrm{minimize}}_{#1}}
\newcommand{\mymax}[1]{\mathop{\textrm{maximize}}_{#1}}
\newcommand{\mymins}[1]{\mathop{\textrm{min.}}_{#1}}
\newcommand{\mymaxs}[1]{\mathop{\textrm{max.}}_{#1}}
\newcommand{\myargmin}[1]{\mathop{\textrm{argmin}}_{#1}}
\newcommand{\myargmax}[1]{\mathop{\textrm{argmax}}_{#1}}
\newcommand{\myst}{\textrm{s.t. }}
\newcommand{\denselist}{\itemsep -1pt}
\newcommand{\sparselist}{\itemsep 1pt}

\newcommand{\cyan}[1]{\textcolor{cyan}{#1}}
\newcommand{\blue}[1]{\textcolor{blue}{#1}}
\newcommand{\magenta}[1]{\textcolor{magenta}{#1}}
\newcommand{\pink}[1]{\textcolor{pink}{#1}}
\newcommand{\green}[1]{\textcolor{green}{#1}}
\newcommand{\gray}[1]{\textcolor{gray}{#1}}
\newcommand{\mygreen}[1]{\textcolor{mygreen}{#1}}
\newcommand{\purple}[1]{\textcolor{purple}{#1}}
\newcommand{\greena}[1]{\textcolor{greena}{#1}}
\newcommand{\bluea}[1]{\textcolor{bluea}{#1}}
\newcommand{\reda}[1]{\textcolor{reda}{#1}}
\def\changemargin#1#2{\list{}{\rightmargin#2\leftmargin#1}\item[]}
\let\endchangemargin=\endlist
\newcommand{\cm}[1]{}
\newcommand{\mhoai}[1]{{\color{magenta}\textbf{[MH: #1]}}}
\newcommand{\mtodo}[1]{{\color{red}$\blacksquare$\textbf{[TODO: #1]}}}
\newcommand{\myheading}[1]{\vspace{1ex}\noindent \textbf{#1}}
\newcommand{\htimesw}[2]{\mbox{$#1$$\times$$#2$}}
\newif\ifshowsolution
\showsolutiontrue
\ifshowsolution
\newcommand{\Comment}[1]{\paragraph{\bf $\bigstar $ COMMENT:} {\sf #1} \bigskip}
\newcommand{\Solution}[2]{\paragraph{\bf $\bigstar $ SOLUTION:} {\sf #2} }
\newcommand{\Mistake}[2]{\paragraph{\bf $\blacksquare$ COMMON MISTAKE #1:} {\sf #2} \bigskip}
\else
\newcommand{\Solution}[2]{\vspace{#1}}
\fi
\newcommand{\truefalse}{
\begin{enumerate}
	\item True
	\item False
\end{enumerate}
}
\newcommand{\yesno}{
\begin{enumerate}
	\item Yes
	\item No
\end{enumerate}
}
\newcommand{\Sref}[1]{Sec.~\ref{#1}}
\newcommand{\Eref}[1]{Eq.~(\ref{#1})}
\newcommand{\Fref}[1]{Fig.~\ref{#1}}
\newcommand{\Tref}[1]{Table~\ref{#1}}
\def\model{LiftRefine}

\maketitle
\begin{strip}
    \centering
    \includegraphics[width=\linewidth]{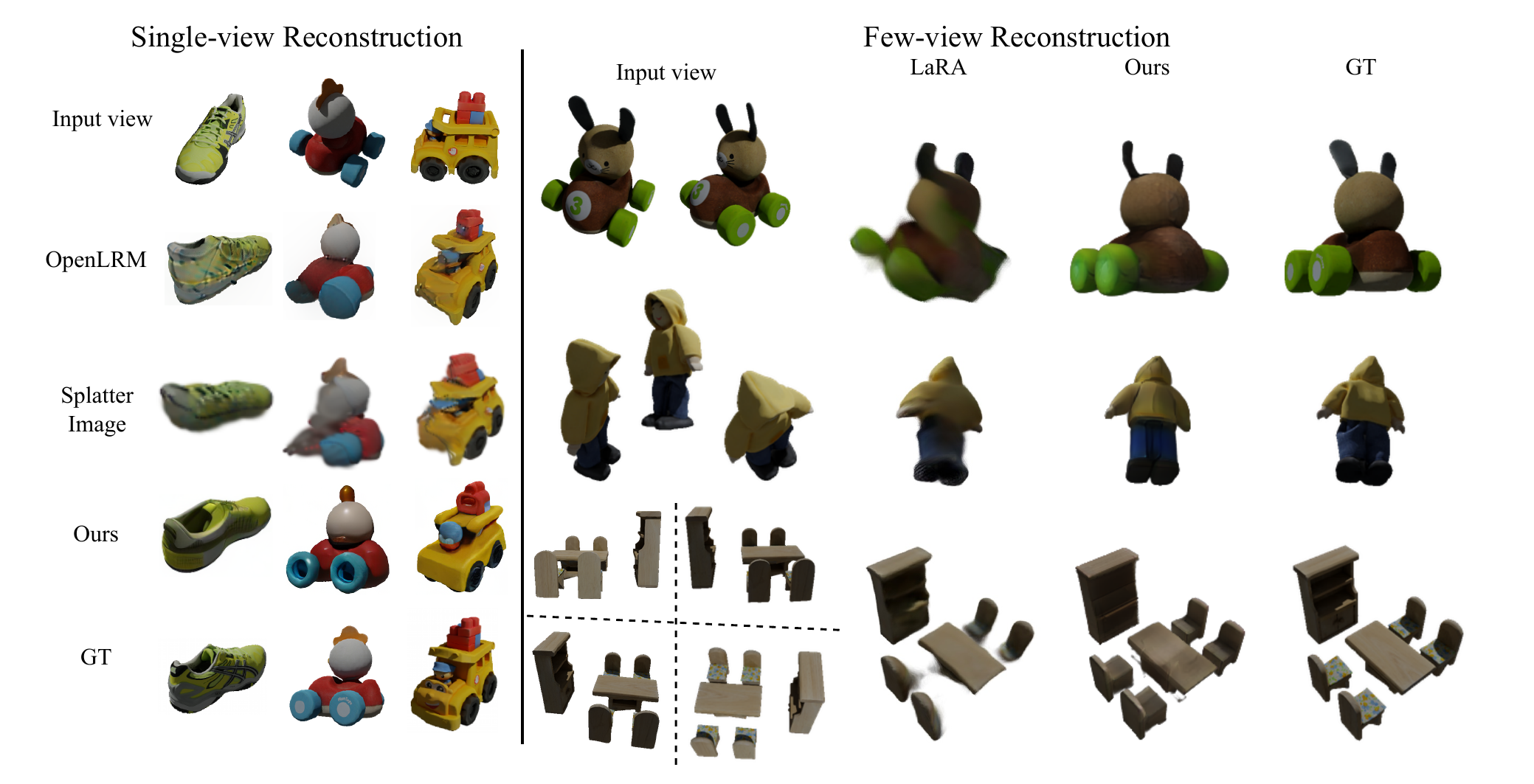}
    \captionof{figure}{Our novel view synthesis addresses both single-view and few-view setting with high-quality reconstruction and rendering. Previous methods like OpenLRM \cite{openlrm} and SplatterImage \cite{szymanowicz23splatter} struggle to accurately reconstruct occluded regions, whereas our method can generate plausible results. For few-view reconstruction, LaRa \cite{LaRa} experiences a rapid decline in performance as the number of input views decreases. In contrast, our method consistently delivers faithful reconstructions across a wide range of input views.}
    \label{fig:teaser}
\end{strip}

\begin{abstract} 

We propose a new view synthesis method via synthesizing a 3D neural field from both single or few-view input images. To address the ill-posed nature of the image-to-3D generation problem, we devise a two-stage method that involves a reconstruction model and a diffusion model for view synthesis. Our reconstruction model first lifts one or more input images to the 3D space from a volume as the coarse-scale 3D representation followed by a tri-plane as the fine-scale 3D representation.
To mitigate the ambiguity in occluded regions, our diffusion model then hallucinates missing details in the rendered images from tri-planes. 
We then introduce a new progressive refinement technique that iteratively applies the reconstruction and diffusion model to gradually synthesize novel views, boosting the overall quality of the 3D representations and their rendering. 
Empirical evaluation demonstrates the superiority of our method over state-of-the-art methods on the synthetic SRN-Car dataset, the in-the-wild CO3D dataset, and large-scale Objaverse dataset while achieving both sampling efficacy and multi-view consistency.

\end{abstract}

\section{Introduction}
View synthesis is a traditional task in computer vision and graphics with typical applications to enhance audience experience in entertainment and telepresence. 
At its core, view synthesis can be solved via image-based rendering and 3D reconstruction methods. 
The recent introduction of diffusion models~\cite{ho2020denoising} and neural radiance field (NeRF) \cite{mildenhall2020nerf} have enabled high-quality image-based rendering and 3D reconstruction, renewing interest in effective and efficient view synthesis.

State-of-the-art view synthesis methods based on neural representations \cite{single_view_mpi, yu2021pixelnerf, wang2021ibrnet, mvsnerf, mildenhall2020nerf, sitzmann2019srns, lin2023visionnerf} tend to exhibit a mean-seeking behavior that often results in blurriness in unseen regions. Inspired by the advance of generative modeling, to address this limitation, several methods incorporate a generative model to synthesize details for the occluded regions. This approach can be tracked with two notable directions: image-based and 3D-based synthesis.

Particularly, image-based synthesis approach involves training an image diffusion model on the 2D view distribution. This method offers numerous advantages, such as directly leveraging prior knowledge from existing large pretrained diffusion models (e.g., Zero123 \cite{liu2023zero1to3} finetuned on Stable Diffusion \cite{rombach2021highresolution}). However, a notable challenge is that these methods cannot guarantee multi-view consistency among the generated images. Typically, they are coupled with auto-regressive sampling techniques or resort to test-time optimization \cite{poole2022dreamfusion, wang2023prolificdreamer}, incurring a significant time cost.

In contrast to image-based models, 3D-based synthesis involves training a 3D diffusion model that ensures multi-view consistency as it directly predicts a 3D representation from the input views. However, training a 3D diffusion model requires substantial memory and the availability of 3D datasets remains limited in scale and diversity compared to massive 2D datasets such as LAION-5B \cite{schuhmann2022laion5b}. These challenges hinder the scalability of this approach.

In this paper, we introduce a two-stage method that takes advantage of both 3D reconstruction and image synthesis for realistic novel view synthesis. 
In the first stage, we propose a novel reconstruction model that integrates both volume and tri-plane features. The motivation behind our fusion comes from two key considerations. Firstly, volumetric representation has demonstrated impressive results in various reconstruction models \cite{chan2023genvs, szymanowicz23viewset_diffusion, karnewar2023holodiffusion}, but its memory-intensive nature limits scalability. Conversely, tri-plane representation offers a more compact alternative that supports higher resolutions. However, existing approaches \cite{lrm2023, anciukevicius2022renderdiffusion} face challenges in transforming 2D images into 3D tri-planes. Utilizing simplistic methods such as 2D-UNet or transformer-based architectures without geometry guidance often leads to mixed-up features within each tri-plane and an inability to capture high-frequency details of 3D objects.
By combining volumetric and tri-plane representations, our approach effectively leverages the geometric interpretation provided by volumetric representation while enjoying the compactness of triplane representation. This integration enables us to overcome the limitations of each individual representation and enhance the reconstruction capability of our model. 
In the second stage, our method focuses on the image-based diffusion paradigm, leveraging its efficiency and access to extensive literature and massive datasets. Inspired by Latent Diffusion \cite{rombach2021highresolution}, our diffusion model operates in latent space to capture the distribution of novel views. 
More importantly, we devise a progressive inference procedure that iteratively applies both stages to gradually boost the quality of the 3D representation and rendered novel views across the initial and target view angles.

In summary, our contributions can be given as follows:
\begin{itemize} 

\item We introduce a novel view synthesis method that employs a 3D reconstruction stage using both volumetric and tri-plane representations, along with an image synthesis stage based on image diffusion to predict novel views. We devise a training strategy to learn both the reconstruction model and the image diffusion model.

\item We present a progressive inference procedure designed to systematically enhance the image quality of unseen regions by iteratively generating intermediate views from the input view toward the target view.

\item We extensively conduct experiments on 
several datasets to validate our approach's effectiveness, consistently showing substantial improvements over previous state-of-the-art methods, as shown in Fig.\ref{fig:teaser}.

\end{itemize}

\section{Related work}

\myheading{Novel view synthesis} has recently gained renewed attention thanks to the introduction of 
neural radiance fields (NeRFs)~\cite{mildenhall2020nerf}. 
NeRF-based novel view synthesis is based on the concept of regressing a neural field from the input images~\cite{single_view_mpi, yu2021pixelnerf, wang2021ibrnet, mvsnerf, mildenhall2020nerf, sitzmann2019srns, lin2023visionnerf} so that novel views at any camera pose can be rendered from this radiance field. A common issue of these methods is that when rendering occluded parts of the scene, these models tend to generate blurry images, primarily because they are mean estimators \cite{chan2023genvs}. A long series of work \cite{chan2023genvs, karnewar2023holodiffusion, liu2023zero1to3, tewari2023forwarddiffusion, szymanowicz23viewset_diffusion, ssdnerf, rombach2021geometryfree, ren2022look, infinite_nature_2020, kim2023nfldm} focus on the use of generative models to capture the underlying distribution of the data in order to generate plausible content for unseen regions. For example, diffusion models \cite{liu2023zero1to3, poseguideddiffusion, zhou2023sparsefusion, watson2022novel, tewari2023forwarddiffusion, szymanowicz23viewset_diffusion, ssdnerf, kim2023nfldm} have demonstrated the ability to synthesize realistic novel views when conditioned on an input image. \cite{kim2023nfldm} uses a two-stage design by first learning a 3D reconstruction model on a 2D dataset and then training a 3D diffusion model. \cite{chan2023genvs, zhou2023sparsefusion, poseguideddiffusion} focus on a 3D-aware diffusion model to refine the deterministic 2D feature map into novel-view images. In the context of the diffusion-based novel view synthesis model, learning from the 2D distribution of novel view images is favored over the 3D distribution of the dataset due to the use of massive pretrained 2D diffusion models and datasets.

\myheading{2D Novel View Diffusion.} Early research efforts \cite{chan2023genvs, gu2023nerfdiff, poseguideddiffusion, zhou2023sparsefusion, watson2022novel} have been dedicated to capture the distribution of 3D objects by analyzing 2D novel image distributions. Notably, Tewari et al. \cite{tewari2023forwarddiffusion}, despite only generating 2D images, have managed to synthesize outputs that closely mimic the 3D distribution, resulting in consistent and high-fidelity visuals. In a similar vein, GeNVS \cite{chan2023genvs} leverages a volumetric representation of the input camera's frustum to create a novel view feature map through volume rendering, which then is used as the condition for the diffusion model. Additionally, certain studies \cite{poseguideddiffusion, zhou2023sparsefusion} employ epipolar lines to better guide the diffusion process. Whereas, another line of work \cite{watson2022novel, liu2023zero1to3} omits explicit 3D geometrical modeling, relying solely on the capabilities of the diffusion model to generate novel 2D views. Despite these advancements, image-based diffusion models still encounter view inconsistency. To mitigate this issue, one ought to to utilize auto-regressive sampling techniques to gradually generate the complete image sequences \cite{chan2023genvs} or to employ test-time optimization with score distillation sampling \cite{gu2023nerfdiff, zhou2023sparsefusion, liu2023zero1to3}, aiming for more accurate and consistent 3D representations.

\myheading{3D Reconstruction Model.} Multi-plane representation \cite{single_view_mpi} and the epipolar line constraint \cite{poseguideddiffusion, zhou2023sparsefusion} are favored methods for novel-view synthesis due to their lightweight nature. However, these methods encounter difficulties when the target novel views are significantly different from the original input views. On the other hand, volumetric representation \cite{chan2023genvs, szymanowicz23viewset_diffusion, karnewar2023holodiffusion} offers impressive accuracy without the above-mentioned limitation, although they suffers from a high memory footprint. To address such a problem, various methods \cite{lrm2023, anciukevicius2022renderdiffusion, ssdnerf} utilize a compact form of 3D volume called tri-plane. However, RenderDiffusion \cite{anciukevicius2022renderdiffusion} adopts a 2D-UNet to encode 2D images and decode them into tri-planes, resulting in mixed-up features within the tri-plane due to the lack of geometry guidance. While LRM \cite{lrm2023} demonstrates impressive performance on large-scale datasets, it still exhibits poor quality on smaller datasets such as SRN-Car.

In our work, \model{} leverages the synergy between a 3D fusion representation and 2D image diffusion to set new standards in novel view synthesis. By addressing the limitations of existing methods in handling 3D scenes, ensuring spatial consistency, and reconstructing fine details in occluded region, our work marks a significant advancement in 3D generation for both single-view and few-view settings.

\section{Proposed Method}

\begin{figure*}[t]
  \centering
  \includegraphics[width=\linewidth ]{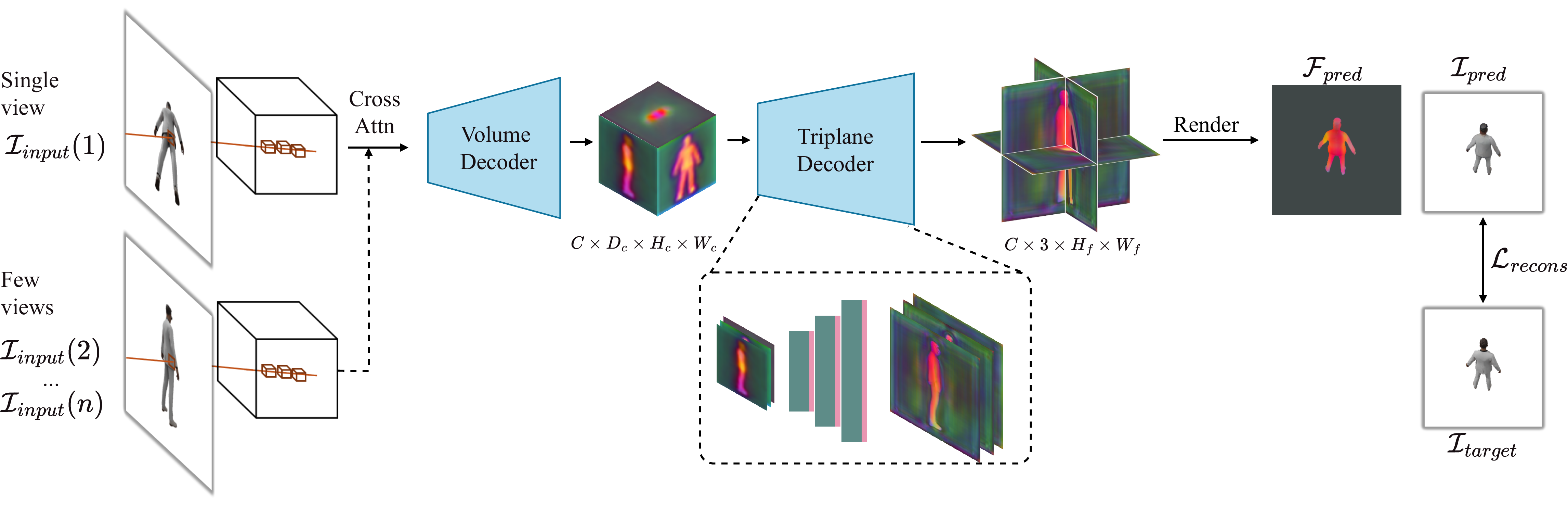}
  \caption{Our Stage 1 involves a reconstruction model to lift the input to 3D representations. Our model supports both single-view and few-view reconstruction, where all input features are aggregated into the volume decoder. The volume is then transformed into a triplane for rendering to novel view images and feature maps.}
  \label{fig:training}
\end{figure*}

We take a two-stage approach for novel view synthesis. Stage 1 involves a reconstruction model that \textbf{lifts} input from a single view or few views to a neural 3D representation, and Stage 2 involves a diffusion model that \textbf{refines} the rendering from the neural representation by hallucinating details for occluded regions. 
The reconstruction model (Section \ref{hybrid_reconstructor}) combines a volumetric radiance field and a tri-plane radiance field for both coarse- and fine-scale 3D representations. Like previous NeRF-based view synthesis methods \cite{mildenhall2020nerf,yu2021pixelnerf,lin2023visionnerf}, our reconstruction model may produce blurry results in occluded regions due to the inherent ambiguity in unseen views. We propose a latent diffusion model conditioned on the novel view feature maps (Section \ref{diffusion_model}) to refine view rendering. 
We propose progressive inference (Section \ref{progressive_enhancement}) that exploits the view consistency of our reconstruction model and the hallucination ability of our diffusion model to effectively enhance the fidelity and completeness of the final reconstructed 3D scene.

\subsection{Lift: 3D Reconstruction}
\label{hybrid_reconstructor}

In the first stage, our reconstruction model transforms the input view(s) into a 3D representation, utilizing both volumetric and tri-plane approaches to generate coarse- and fine-scale representation. An overview of our proposed reconstruction model is shown in Fig.~\ref{fig:training}.

\myheading{Coarse-scale Volumetric Radiance Field.} Given an input image $\mI_{input} \in \R^{C \times H \times W}$ and its corresponding camera pose $\Psi_{input}$, we begin by extracting a feature map using a pretrained feature extractor (e.g., ResNet34, Dino-v2). Next, every voxel coordinate of a unit volume where the center is the world origin, is projected onto the image screen, and bilinear interpolation is applied to obtain the feature for each voxel. 
This process results in a coarse single-view volume feature. 
This coarse feature is then processed by a 3D convolutional encoder, the output of which is then aggregated via a cross-attention mechanism in the volume decoder to output a unified multi-view feature volume, similar to the approach in \cite{szymanowicz23viewset_diffusion}. 
The final output of this stage is a coarse-scale multi-view feature volume $\mV_{low} \in \R^{C \times D_{c} \times H_{c} \times W_{c}}$, where $D_{c}$, $H_{c}$, $W_{c}$ are the coarse-scale volume dimension, respectively.

While the volumetric radiance field can be used to predict a 3D scene, the resulting images or features often suffer from low resolution, leading to visual artifacts such as blurriness and grid-like patterns. Although increasing the resolution of the feature volume could improve image quality, the memory requirements grow cubically $O(n^3)$ with  the volume dimension $n$, quickly leading to out-of-memory issue. To address this problem, we gradually enhance the quality of the reconstructed images by progressively upsampling the low-resolution feature volume $\mV_{low}$ into high-resolution tri-plane $\mT_{high}$. Leveraging tri-plane features allows us to achieve significantly higher resolution rendering compared to volumetric methods, thereby improving the quality of novel views.

\myheading{Fine-scale Tri-plane Radiance Field.}
To reconstruct the tri-plane representation $\mT_{high}$ from a low-resolution feature volume $\mV_{low}$, we first project the feature volume onto three orthogonal planes to form depth-aware feature planes $F_{xy}$, $F_{xz}$ and $F_{yz}$. Each depth-aware feature plane is then fed to Triplane Decoder which consist of  multiple upsampling blocks. Each block include a convolutional network with upsampling layers to refine details and increase the resolution. The final output of three feature planes is reshaped to construct the high-resolution tri-plane radiance field $\mT_{high} \in \R^{C \times 3 \times H_{f} \times W_{f}}$, where $H_{f}$, $W_{f}$ represent the fine-scale image dimensions. This tri-plane representation can then be rendered to generate color images and feature maps:
\begin{align}
\label{eq:highres}
\mI_{pred}, \mF_{pred} = \mR_{tri}(\mT_{high}, \Psi_{target}),
\end{align}
where $\mR_{tri}$ is the function that renders the triplane to image $\mI_{pred}$ and feature map $\mF_{pred}$ at the target camera pose $\Psi_{target}$. The feature map $\mF_{pred}$ will be used as a condition for rendering diffusion in the following stage. To train the reconstruction model, we combine the standard L2 loss and LPIPS loss:
\begin{align}
\label{eq:highresloss}
\mL_{recons} = \| (\mI_{pred} - \mI_{target})\|_2^2 + \lambda~\mathrm{LPIPS}(\mI_{pred}, \mI_{target}).
\end{align}

We discover that our reconstructor excels at generating detailed 3D objects from just a few views. In fact, it performs well in rendering novel views with minimal ambiguity, even without the assistance of a diffusion model. However, when information on occluded regions is lacking, the reconstructor tends to produce blurry results. To overcome this issue, we incorporate a diffusion model, as detailed below.

\begin{figure}[t]
  \centering
  \includegraphics[width=\linewidth ]{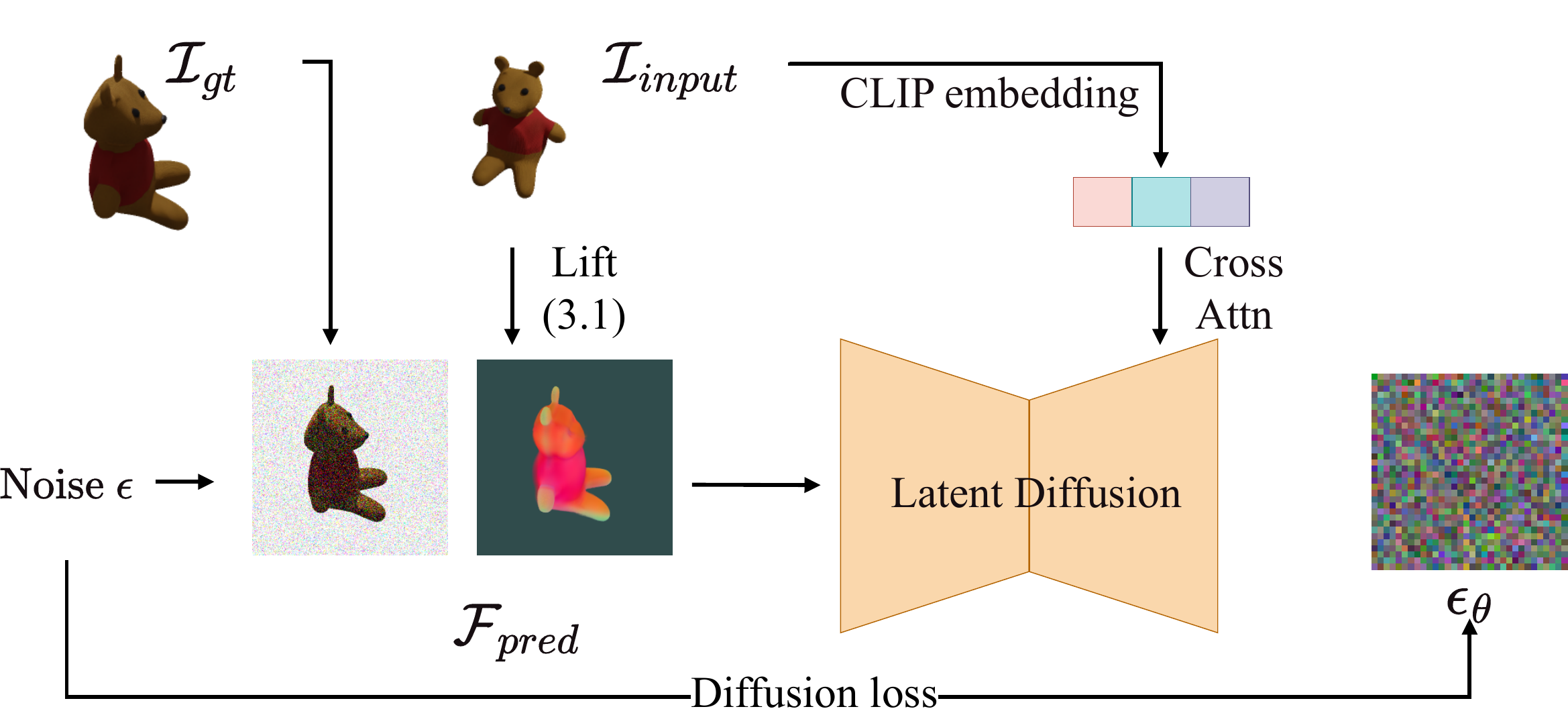}
  \caption{Our Stage 2 involves a conditional rendering diffusion model that aims to refine the rendered novel view from Stage 1 with additional details from a latent diffusion model.}
  \label{fig:diffusion_training}
\end{figure}

\subsection{Refine: Conditional Rendering Diffusion}
\label{diffusion_model}

In the second stage, our method refines blurry novel view images in the previous step using a diffusion model conditioned on the feature map rendered from the first stage. 
This second stage is depicted in Fig.~\ref{fig:diffusion_training}.
Our diffusion model, based on latent diffusion~\cite{rombach2021highresolution}, can be trained as follows. We first precompute a conditional rendering dataset, where each sample contains a pair of input view $\mI_{input}$ and a ground truth novel view $\mI_{gt}$. 
We perform Stage 1 (Lift) to predict the feature map of the novel view $\mF_{pred}$ using Eq.~\ref{eq:highres}. 
The denoising process of our latent diffusion model then follows.
The key idea is to make the diffusion model learn to refine the novel view conditioned by the input view. 
Therefore, for each data sample and time step $t$, we add Gaussian noise $\epsilon \sim \mathcal{N}(0, I)$ to the novel view $\mI_{gt}$, which is then concatenated with the predicted feature map $\mF_{pred}$ to form the input $x_t$ for the denoising U-Net. 
We further extract the CLIP embedding of the input view $\mI_{input}$ and use this embedding to condition the U-Net via cross attentions. The diffusion model predicts the added noise, which is trained by the following diffusion loss: 
\begin{align}
\label{eq:diffusionloss}
\mL_{diffusion} = \mathbb{E}_{t, \epsilon} \| \epsilon_\theta(x_{t}, t \mid \mF_{pred}, \mI_{clip}) - \epsilon \|_2^2,
\end{align}
where $\epsilon_{\theta}$ is the denoising U-Net parameterized by $\theta$. 

\subsection{Progressive Inference}
\label{progressive_enhancement}
\begin{figure}[t]
  \centering
  \includegraphics[width=\linewidth]{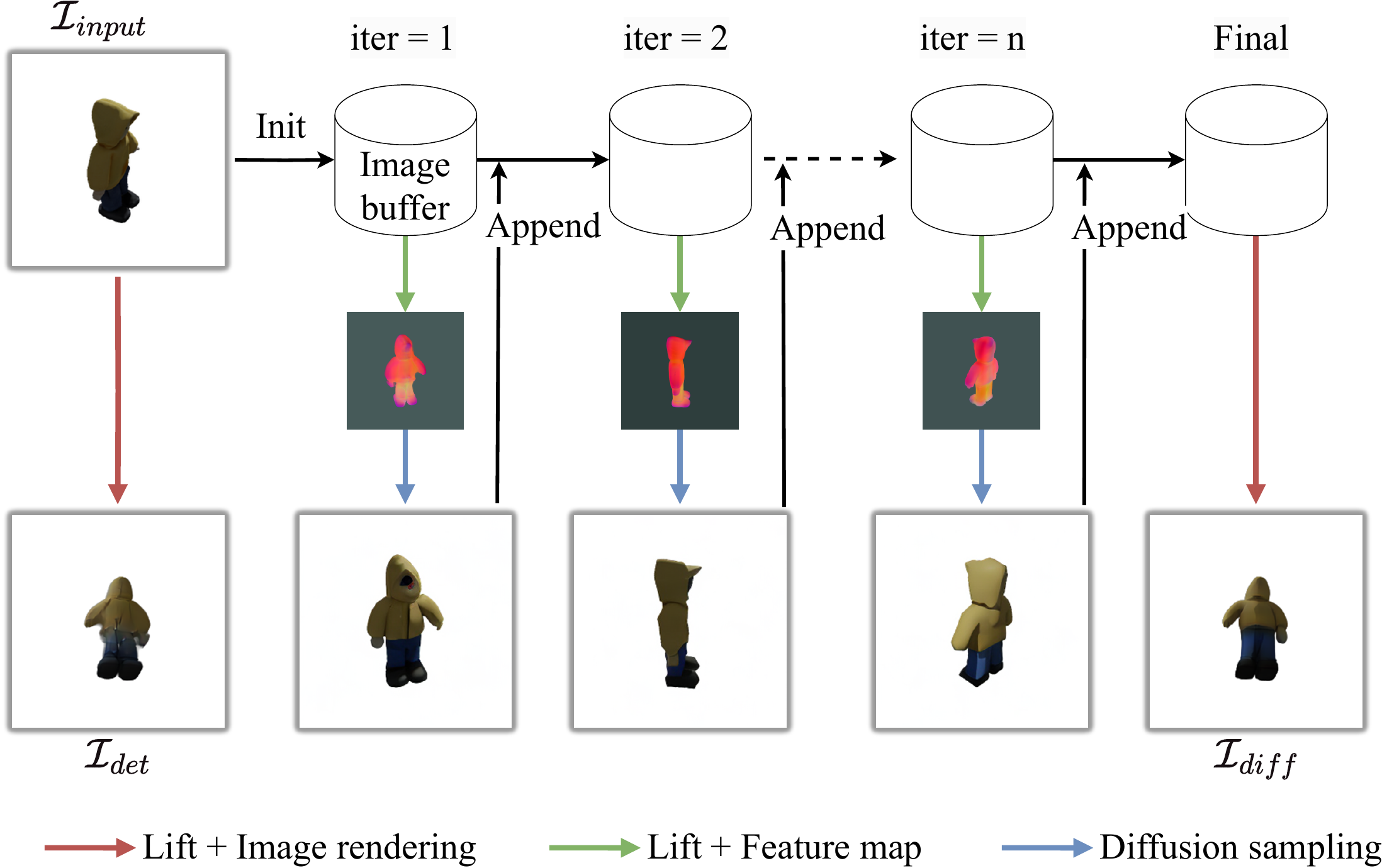}
  \caption{Progressive inference. Our method reconstructs and generates intermediated views, progressively refining the quality of the 3D representation and its rendering.}
  \label{fig:progressive}
\end{figure}

In scenarios with significant ambiguities, the reconstruction model tends to produce blurry images in occluded regions, but albeit with strong view-to-view consistency. Conversely, the diffusion model generates high-quality results but may introduce inconsistencies across views. To leverage the strengths of both models, we introduce a progressive inference procedure that integrates the high-fidelity outputs from the diffusion model into the consistent 3D reconstructor, gradually filling in unseen regions and enhancing the overall reconstruction quality. This technique is demonstrated in Fig.~\ref{fig:progressive}.

Our inference process begins by initializing an image buffer to store all input views. In each iteration of the inference, we render a feature map using the reconstructor, conditioned on all images currently in the buffer. This feature map is then fed to the diffusion model, which generates an intermediate novel view image. The newly generated image is subsequently added to the buffer. All images in the buffer are then used for the next iteration to generate the next intermediate novel view image. At the final iteration, we applied the reconstructor to generate the final novel view image. 
In our experiments, for clarity, we refer to the novel views rendered by the reconstructor without using progressive inference as deterministic novel views $\mI_{det}$, and novel views from progressive inference as $\mI_{diff}$, respectively.

\section{Experiments}

\myheading{Training.} We first train our reconstruction model, followed by precomputing the conditional renderings to train the diffusion model. Our training is conducted on three different datasets: CO3D \cite{reizenstein21co3d}, Objaverse \cite{objaverse}, and Shapenet-SRN Cars \cite{sitzmann2019srns}. The CO3D dataset, an in-the-wild collection, is utilized to assess the robustness of our method in real-world conditions where imperfections are present. The Objaverse dataset demonstrates the scalability of our model on a large-scale dataset. Finally, Shapenet-SRN Cars, a synthetic dataset focused on single-category objects, is employed in our ablation study. More details on ShapeNet-SRN Cars are provided in the supplementary material.

\myheading{Inference.} We evaluate our method with two different settings: deterministic and a diffusion-based. In the deterministic setting, we utilize the reconstruction model to synthesize a tri-plane from the input view, and then render the novel view images \emph{without} progressive refinement. 
In the diffusion-based setting, we employ progressive inference as outlined in Sec.~\ref{progressive_enhancement}, and use the final tri-plane to render the refined novel view. Please prefer to supplementary material for video qualitative results.

\myheading{Dataset and Evaluation Protocol.} 
Some methods are trained for a specific setting, such as single-view or few-views reconstruction. Therefore, we divide our evaluation protocol into these two settings. Notably, our method is effective in both scenarios.

We conduct both single-view and few-views evaluations on the CO3D dataset \cite{reizenstein21co3d} and the Google Scanned Object (GSO) dataset \cite{gso}.
 We assess our method at a resolution of 128x128 for in-the-wild CO3D dataset \cite{reizenstein21co3d} and 256x256 for the GSO dataset \cite{gso}.

\subsection{CO3D}
On the CO3D dataset, we conduct a  comparative analysis on four classes (Hydrant, Teddybear, Vase and Plant) on two state-of-the-art methods: ViewsetDiffusion \cite{szymanowicz23viewset_diffusion} and SparseFusion \cite{zhou2023sparsefusion}. The quantitative results are summarized in Tab.~\ref{tab:main_co3d}. 
In both single-view and few-views reconstruction tasks, our model consistently surpasses previous methods across key metrics, including PSNR, SSIM, and LPIPS, highlighting its effectiveness in capturing fine details and preserving image quality. In the diffusion setting, while there is a decrease in pixel-wise metrics, our approach shows improvements in object quality, evidenced by better SSIM and LPIPS scores. Additionally, for the distribution-based metric FID, our diffusion-based setting demonstrates significant gains over the deterministic setting, indicating the efficacy of progressive inference in enhancing the overall quality of the generated images. It is important to note that SpareFusion generates images, while ViewsetDiffusion and our method reconstruct 3D representations. Consequently, SpareFusion exhibits a higher FID but lower consistency compared to our method.

\begin{figure}
  \centering
  \includegraphics[width=1.0\linewidth]{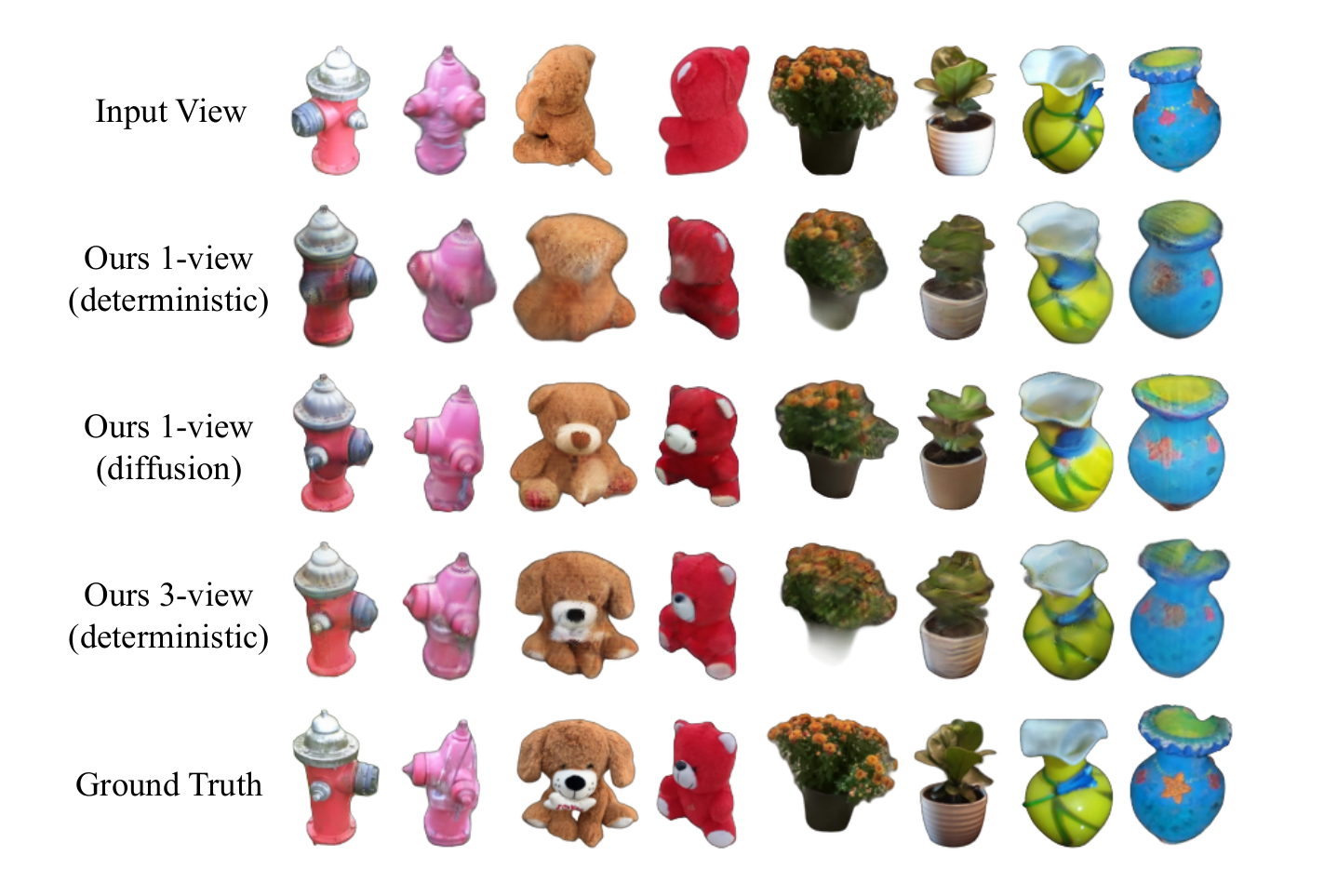}
  \caption{Qualitative results on CO3D Dataset.}
  \label{fig:main_co3d}
\end{figure}

In Fig.~\ref{fig:main_co3d}, for a deterministic setting with a single input view, we observe high-quality synthesis in visible regions but encounter blurriness in occluded areas. Upon employing progressive inference, the final object fidelity is significantly improved, approaching the performance achieved in the deterministic setting with three input views. For results on multi-view consistency, please refer to the supplementary material, where a video presentation is provided.

\subsection{Google Scanned Object} 
On the GSO dataset, we compare our method under the single-view setting with OpenLRM \cite{openlrm}, an open-source version of the LRM model~\cite{lrm2023}, and Splatter Image \cite{szymanowicz23splatter}, as these methods are designed for single-view reconstruction.
Results are shown in Tab.~\ref{tab:main_gso_256} and Fig.~\ref{fig:teaser}.
Our model outperforms previous methods for more than 1 db in PSNR while having competitive performance on SSIM and LPIPS. Compared to LRM which is purely transformer-based, our method effectively leverages inductive bias from the volume and triplane representations, leading to more robust models with reduced training time. Our method requires 4 NVIDIA A100 GPUs for training in 7 days while LRM needs 128 A100 GPUs for training in 3 days.

In the few-view reconstruction task, we compare our model with the current SOTA method LaRa \cite{LaRa}.
The results is shown in Tab.~\ref{tab:main_gso_256}. Our model outperforms LaRa significantly when using 2 or 3 input views and remains competitive with 4 input views. This advantage arises from our volume-based representation, which excels at reconstructing unseen regions. In contrast, LaRa's Gaussian splatting method struggles to render occluded areas when input views are limited.

\begin{table}[t]
\centering
\resizebox{\linewidth}{!}{
\begin{tabular}{lcccl}
\toprule
& & \multicolumn{3}{c}{1-view}   \\
\cmidrule(lr){2-5}
&  PSNR$\uparrow$ & SSIM$\uparrow$ & LPIPS$\downarrow$  & FID$\downarrow$  \\ 
SparseFusion   & 16.45 & 0.652 & 0.278 & \underline{46.5} \\

ViewsetDiffusion &  18.41 & 0.684 & 0.280 & 99.6\\

Ours (Det) & \textbf{20.38} & \textbf{0.747} & \underline{0.204} & 73.4 \\

Ours (Diff) & \underline{20.10} & \underline{0.744} & \textbf{0.195} & \textbf{39.6} \\

\midrule
& & \multicolumn{3}{c}{3-view} \\
\cmidrule(lr){2-5}
& PSNR$\uparrow$ & SSIM$\uparrow$ & LPIPS$\downarrow$  & FID$\downarrow$  \\ 
SparseFusion   & 21.48 & 0.773 & 0.175 & \textbf{29.3} \\

ViewsetDiffusion &  21.86 & 0.752 & 0.241 & 91.7\\

Ours (Det) &  \textbf{22.82} & \textbf{0.800} & \underline{0.166} & 59.5 \\
Ours (Diff) &  \underline{22.66} & \underline{0.797} & \textbf{0.161} & \underline{43.3} \\

\bottomrule

\end{tabular}}
\centering
\caption{Results on CO3D dataset for single and few-view reconstruction.}
\label{tab:main_co3d}
\end{table}

\begin{table}
    \centering
    \resizebox{\linewidth}{!}{
\begin{tabular}{clccc}
    \toprule
    \# Views & Method & PSNR $\uparrow$ & SSIM $\uparrow$ & LPIPS $\downarrow$ \\
    \midrule
     & OpenLRM & 18.06 & 0.840 & 0.129 \\
    1 & Splatter Image & 21.06 & 0.880 & \textbf{0.111} \\
    & Ours (Det) & 22.23 & 0.880 & 0.113 \\
    & Ours (Diff) & \textbf{22.55} & \textbf{0.895} & 0.116 \\
    \midrule
     & LaRa & 19.59 & 0.877 & 0.151 \\
    2 & Ours (Det) & 23.82 & 0.908 & 0.092 \\
     & Ours (Diff) &\textbf{23.91} & \textbf{0.909} & \textbf{0.090} \\
    \midrule
     & LaRa & 23.92 & 0.915 & 0.112 \\
    3 & Ours (Det) & 24.99 & 0.916 & 0.085 \\
     & Ours (Diff) &\textbf{25.11} & \textbf{0.917} & \textbf{0.083} \\
    \midrule
     & LaRa & \textbf{26.03} & \textbf{0.930} & 0.098 \\
    4 & Ours (Det) & 25.64 & 0.920 & 0.082 \\
     & Ours (Diff) & 25.79 & 0.922 & \textbf{0.080} \\
    \bottomrule
\end{tabular}}
\caption{Results on GSO dataset for single and few-view reconstruction.}
    \label{tab:main_gso_256}
\end{table}

\subsection{Ablation Study}
In this section, we validate the impact of our design choices. Unless otherwise mentioned, we drop the 2D conditional diffusion model and perform the ablation studies with the deterministic setting.

\myheading{Analysis of 3D Representations.} To demonstrate the efficacy of incorporating low-resolution volume and high-resolution tri-plane representations, we conducted several experiments on CO3D-Hydrant, as shown in Tab.~\ref{tab:ablate_co3d}. Our method is denoted as the default setting. Firstly, by replacing our upsampler layers with bicubic interpolation (setting C), we observed a significant drop in the performance of the reconstructor, nearly 1dB. This indicates the importance of our upsampler in preserving image quality and details during reconstruction. Additionally, we replaced the low-resolution volume features with low-resolution tri-plane representations (setting B). We achieved this by projecting the volume features, obtained after lifting from the input view, onto a tri-plane, and then replacing every 3D layer in the 3D encoder and decoder with 2D convolutional layers to operate on tri-plane features. 
However, the results showed that the performance of setting B is inferior to using volume features as the low-resolution representation. We hypothesize that this is because working on volume at low-resolution allows for more informative 3D features rather than its compact tri-plane version. 

\begin{table}[t]
\centering
    \begin{tabular}{lccc}
        \toprule
        Method & PSNR $\uparrow$ & SSIM $\uparrow$ & LPIPS $\downarrow$ \\
        \midrule
    Default & \textbf{24.56} & \textbf{0.868} & \textbf{0.100} \\
       B & 24.06 & 0.860 & 0.113 \\
       C & 23.66 & 0.853 & 0.111 \\

        \bottomrule
    \end{tabular}
    \caption{Experiments with the impact of three different configurations for the 3D reconstructor on CO3D dataset}
    \label{tab:ablate_co3d}
\end{table}

\begin{table}[t]
    \centering
    \begin{tabular}{lccc}
        \toprule
        Method & PSNR $\uparrow$ & SSIM $\uparrow$ & LPIPS $\downarrow$ \\
        \midrule
        OpenLRM & 22.88 & 0.910 & 0.082 \\
        Triplane  & 21.83 & 0.898 & 0.087 \\
        Volume & 22.78 & 0.908 & 0.087 \\
        Ours & \textbf{23.76} & \textbf{0.921} & \textbf{0.067} \\
        \bottomrule
    \end{tabular}
    \caption{Quantitative results with different image-to-triplane architecture on ShapeNet-SRN Cars dataset.}
    \label{tab:ablate_backbone}
\end{table}

\begin{figure}[t]
  \centering
      \centering
      \includegraphics[width=0.7\linewidth]{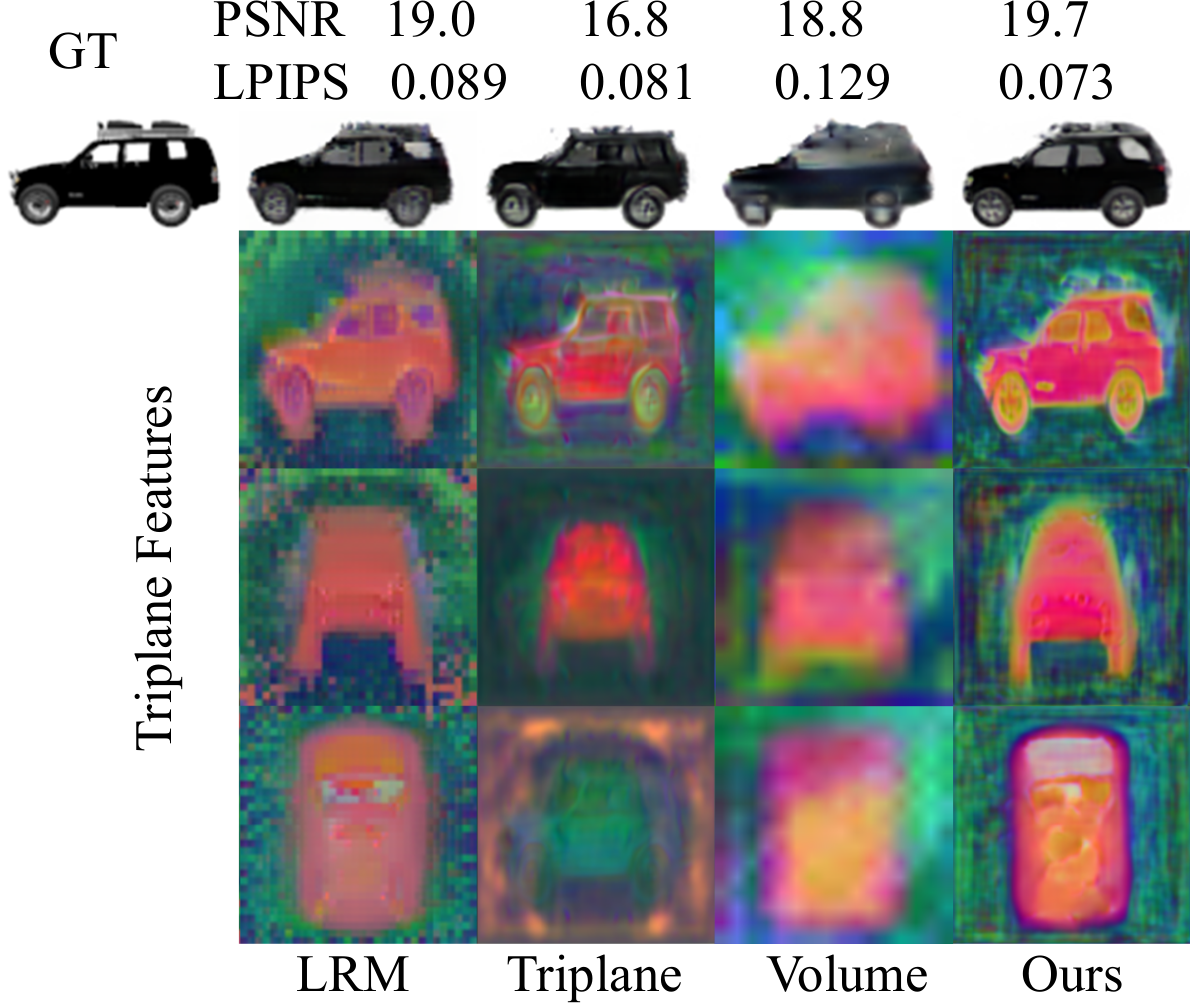}
      \caption{Feature maps of a ShapeNet-SRN Car example with different image-to-triplane backbones.}
      \label{fig:ablate_backbone}
  \end{figure}

\myheading{Analysis of Image-to-Triplane Backbone.}
To study the importance of the low-resolution volume, we conducted a series of experiments with different image-to-triplane backbones. 
The quantitative results are presented in Tab.~\ref{tab:ablate_backbone}.
Firstly, we utilized LRM \cite{lrm2023}, a transformer-based reconstruction model renowned for its ability to output tri-plane representations from single images. While LRM demonstrated remarkable performance when applied to large datasets, its efficacy diminishes when working on smaller datasets such as ShapeNet-SRN Cars. This is because the transformers are unaware of 3D inductive bias, and instead are tasked to implicitly learn spatial relationship, which deteriorates its performance when training on a small dataset like ShapeNet. Notably in Fig.~\ref{fig:ablate_backbone}, the tri-plane features from LRM exhibited pixelated noise, thereby impairing its overall performance.
Subsequently, following \cite{anciukevicius2022renderdiffusion}, we leveraged a 2D UNet to encode and decode 2D images into 3D tri-planes without volume representation. However, we encountered a significant challenge as the output tri-plane appears mixed up due to the absence of channel decoupling during the encoding and decoding stages of the 2D UNet. Consequently, this led to a notable drop in the PSNR metric, indicating a degradation in the quality of reconstruction.
Lastly, we modify our backbone to use only the volume representation. The results were blurry due to the low volume resolution, resulting in high PSNR but low LPIPS scores. 

\myheading{Study on Progressive Inference.}
We investigated the effects of increasing the number of iterations during progressive inference. In Tab.~\ref{tab:ablate_interpolation}, while we observed a decline in pixel-wise metrics such as PSNR and SSIM, there was a clear improvement in the semantic quality of the generated images, reflected by metrics like FID.
We found that using 4 iterations leads to the best balanced result across metrics. Therefore, we used 4 iterations for progressive refinement in our experiments.

\begin{figure}[t]
  \centering
      \centering
      \includegraphics[width=\linewidth]{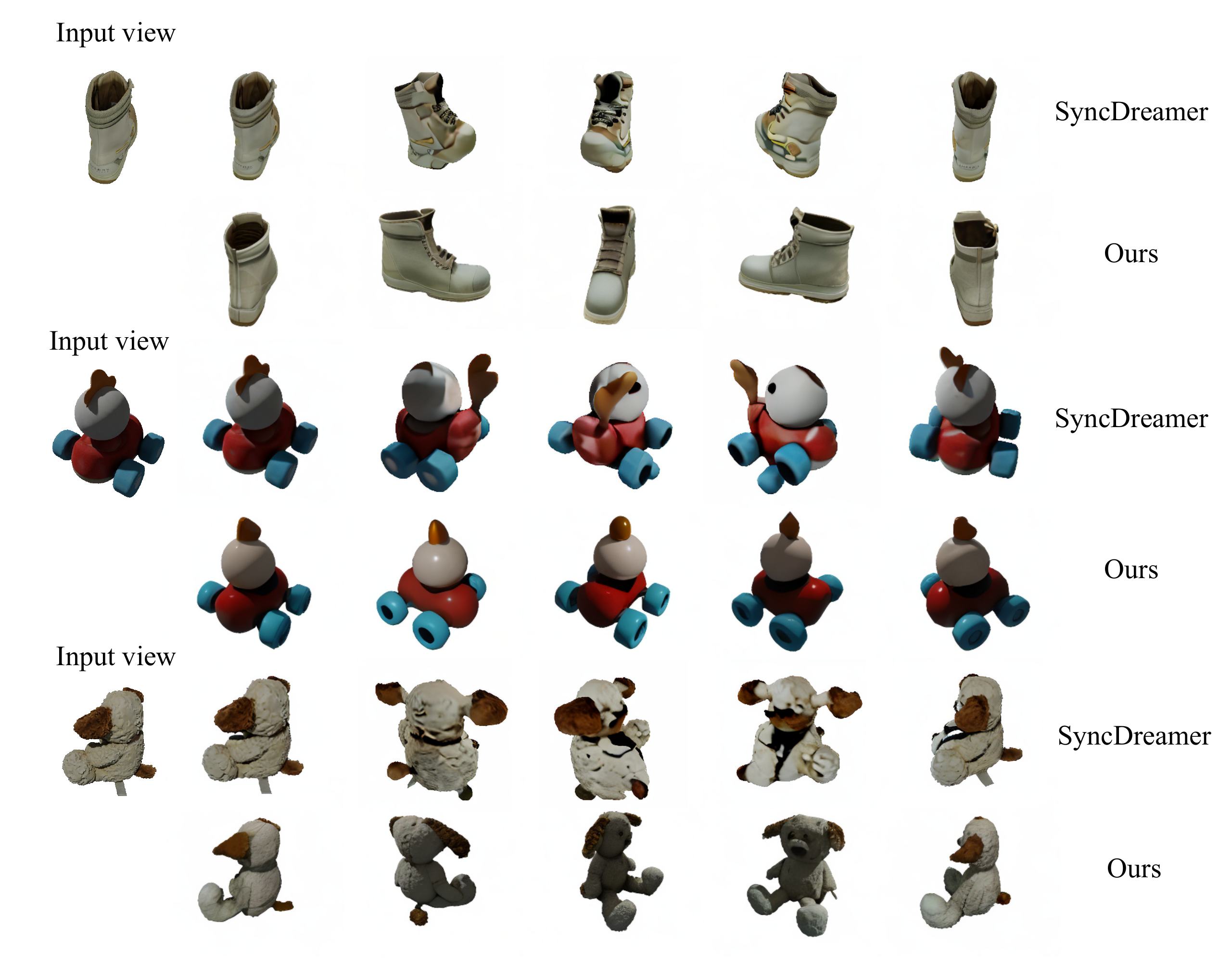}
      \caption{Qualitative comparision with SyncDreamer.}
      \label{fig:syncdreamer}
  \end{figure}

\myheading{Comparision with Multi-view Generation.}
To position our method against SOTA methods in multi-view generation~\cite{shi2024MVDream,wang2023imagedream,long2024wonder3d,liu2023syncdreamer,xu2024dmv3d}, we compare our method with SyncDreamer \cite{liu2023syncdreamer}, a multi-view diffusion model that generates 16 views from input images. We found that although SyncDreamer demonstrates impressive view consistency, their method only works well for input views at frontal angles. Their performance significantly deteriorates when synthesizing from side or rear views of objects. In contrast, our method consistently maintains high-fidelity images across all perspectives as shown in Fig.~\ref{fig:syncdreamer}. For a quantitative comparison with SyncDreamer on novel view synthesis, please refer to the supplementary material.

\begin{table}
    \centering
        \begin{tabular}{cccccc}
            \toprule
            \# iters & PSNR $\uparrow$ & SSIM $\downarrow$ & LPIPS $\downarrow$ & FID $\downarrow$ \\
            \midrule 
            0      & \textbf{22.64} & 0.834 & 0.118 & 42.7  \\
            1      & 22.61 & \textbf{0.845} & \textbf{0.105} & 37.0   \\
            
            2      & 22.43 & 0.843 & 0.107 & 35.4  \\
            
            4      & 22.12 & 0.840 & 0.108 & 34.7 \\
            
            6      & 21.89 & 0.836 & 0.111 & \textbf{34.0}  \\
            
            8     & 21.73 & 0.834 & 0.113 & 34.7  \\
            \bottomrule
        \end{tabular}  
    \caption{The impact of the number of interpolated views on the single-view reconstruction task with CO3D-Hydrant.}
    \label{tab:ablate_interpolation}
\end{table}

\section{Discussion and Conclusion}
In conclusion, this paper introduces \model{}, a new method for novel view synthesis combining 3D reconstruction with image-based diffusion empowerd by a progressive refinement procedure. Our two-stage method achieves SOTA results, delivering realistic and consistent novel views. Extensive testing on diverse datasets validates its effectiveness. 

While our method produces plausible results across datasets, we found that our method struggles in some cases such as the Plant category in CO3D, where the synthesized novel views tend to be blurry. This could be because of the high-resolution details in the plant images that are not well captured by the current resolution of our representations. Improving the high-frequency rendering of this category would be our future work.

\bibliography{aaai25}

\maketitlesupplementary
\begin{abstract}
    
In this supplementary document, we first provide more details about our architecture design in Sec.~\ref{sec:architecture}, and then discuss training and inference details in Sec.~\ref{sec:training_inference}. 
We then present additional experiments in Sec.~\ref{sec:additional_experiment} with more qualitative results in Sec.~\ref{sec:qualitative}. Readers are encouraged to view the supplementary videos for more visual results of our method.

\end{abstract}

\section{Architecture}
\label{sec:architecture}
\subsection{Reconstructor}
Our reconstructor takes in $N$ input frames and outputs a single feature tri-plane representation that can be used to render arbitrary novel views.

\myheading{Encoder.} Our encoder is based on ViewsetDiffusion \cite{szymanowicz23viewset_diffusion} but we replace the feature extractor with a pretrained ResNet34 \cite{he2015deep} or Dino-V2 for enhanced feature extraction capabilities. We adjust the output of the reconstructor, changing from a radiance field volume to a feature volume by increasing the channel dimension from 4 to 32.

\begin{figure}[t]
  \centering
  \includegraphics[width=\linewidth]{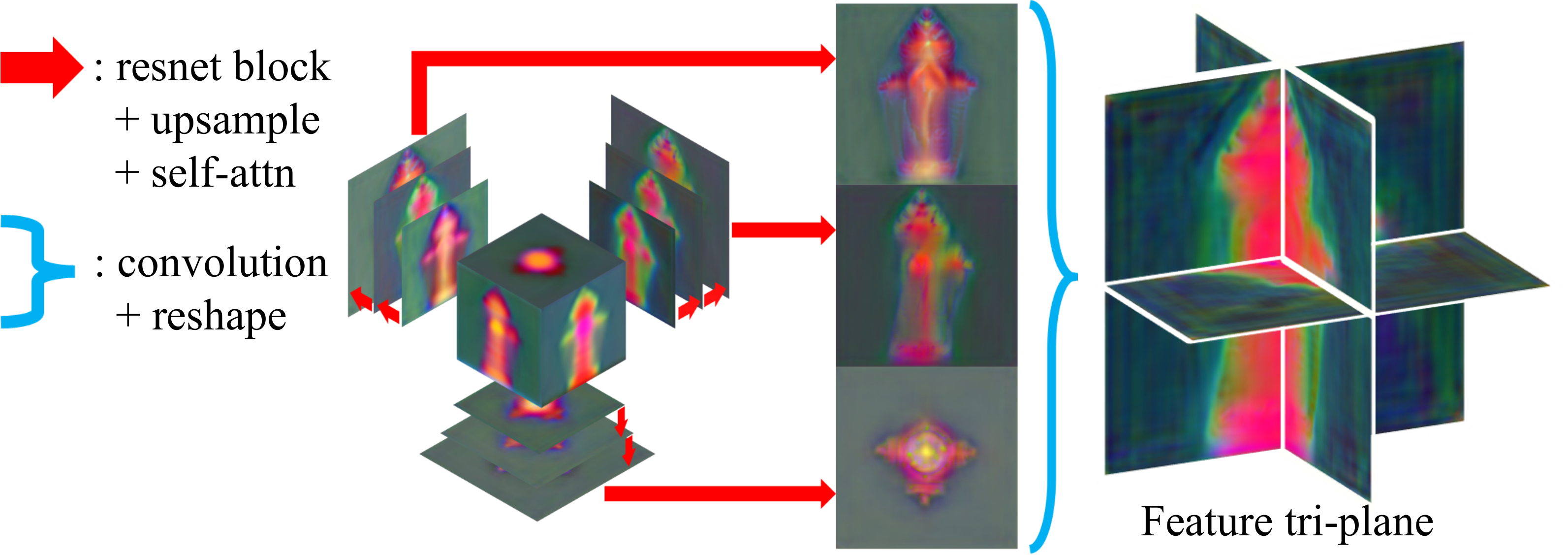}
  \caption{The tri-plane decoder architecture takes a feature volume as input. This feature is then projected into three orthogonal planes to form depth-aware feature planes. Each of these feature planes undergoes processing through multiple upsampler blocks, illustrated by the red arrows. Finally, the feature planes are passed through a final convolution layer and reshaped (depicted by a blue bracket) to form a feature tri-plane.}
    \label{fig:triplane_deocder}
\end{figure}

\myheading{Decoder.} Our tri-plane decoder architecture is based on a 2D UNet decoder~\cite{ho2020denoising}, comprising multiple upsampler blocks. Each upsampler block consists of a ResNet block, an upsample layer, and a self-attention layer at the end. Further details on our tri-plane decoder are illustrated in Fig.~\ref{fig:triplane_deocder}. In our experiments, we upscale the resolution of the feature volume, initially sized at 32, to a feature tri-plane with resolution of 256.

\subsection{Diffusion model} 
Our diffusion operates in the latent space, where we use a VAE model to map an input image $\mI \in \mathbb{R}^{3 \times 256 \times 256}$ to a latent representation  $z \in \mathbb{R}^{4 \times 32 \times 32}$. We use the UNet architecture from Zero123 \cite{liu2023zero1to3} for our 2D conditional diffusion model and increase the condition channel of the original Zero123 model from 4 to 32 to match our conditional feature map. Additionally, we leverage their pretrained model as an initialization for our training process.

\section{Implementation Details}
\label{sec:training_inference}
\myheading{Training.} In the first stage, we train the reconstructor until the validation accuracy stall. For data sampling strategy, we randomly sample 1 to 3 image(s) as input $\mI_{input}$ and 1 image as $\mI_{target}$. After that, we precompute the conditional renderings and follow the same data sampling strategy as training reconstructor to train the diffusion model. 

\myheading{Inference.} We use progressive inference (Fig. 4 in our main paper) to combine our 3D reconstructor and our 2D diffusion model. For CO3D dataset, we interpolate $n$ camera poses between the input and target camera pose. For GSO dataset, we calculate the azimuth and elevation of input view and interpolate $n$ camera poses on the spherical trajectory. In our main experiment, we employ 200-steps DDIM sampling \cite{song2022denoising} with classifier free guidance $=2.0$ to sample the images.

\myheading{Optimization.} We employ the Adam optimizer \cite{kingma2017adam} and cosine scheduler with a learning rate of $1 \times 10^{-6}$ for our reconstructor and $1 \times 10^{-7}$ for our diffusion model. We set the maximum 300k training steps and but stop the training if there is no improvement. The total batch size is 128 and 256 across 4 A100 GPUs for reconstructor and diffusion model respectively. In $\mL_{recons}$, we set $\lambda$ to 0.1 for all experiments.

\section{Additional Experiments}
\label{sec:additional_experiment}

\begin{figure}[t]
  \centering
      \centering
      \includegraphics[width=\linewidth]{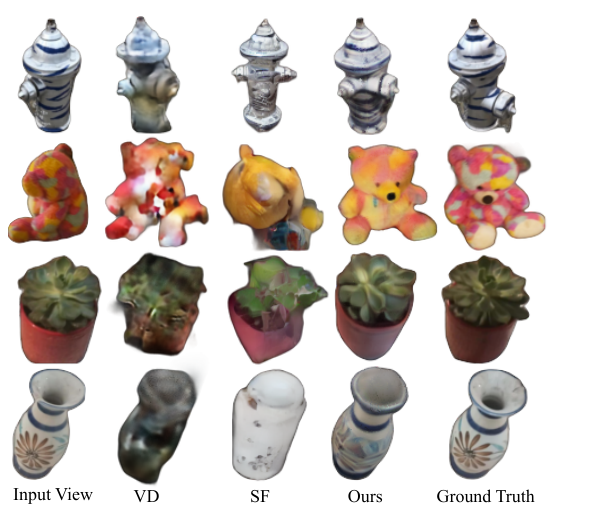}
      \caption{Novel view synthesis from a single input image on CO3D dataset.}
      \label{fig:co3d_comparison}
  \end{figure}

\myheading{Results on ShapeNet-SRN Cars.} 
We evaluate our method at a resolution of 128x128 for the single-view reconstruction on ShapeNet-SRN Cars dataset \cite{sitzmann2019srns}. Following the evaluation protocol from PixelNerf \cite{yu2021pixelnerf}, we adopt their train/val/test split and utilize the $64^{th}$ view as input, while the remaining 250 views serve as unseen target views. 
For Shapenet SRN-Car, we employ the pretrained models from the baselines for inference whenever available. Otherwise, we utilize their reported results for our quantitative analysis.  

\begin{figure}[t]
  \centering
  \includegraphics[width=\linewidth]{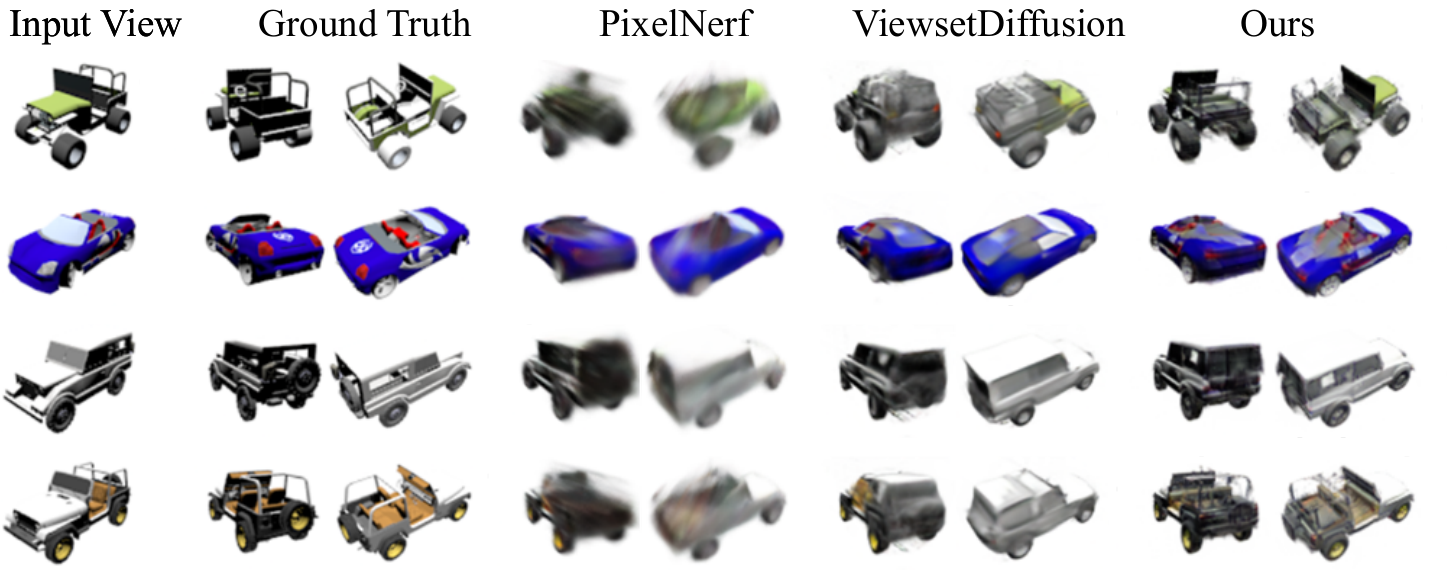}
  \caption{Qualitative results on SRN-Cars.}
  \label{fig:main_shapenet}
\end{figure}

In Tab.~\ref{tab:main_shapenet}, we provide a comprehensive evaluation of our method in comparison to SOTA methods. In the deterministic setting, our model demonstrates superior quantitative performance, notably surpassing regression-based models such as PixelNeRF \cite{yu2021pixelnerf} and VisionNeRF \cite{lin2023visionnerf} across pixel-wise metrics like PSNR and SSIM, as well as perceptual metric LPIPS. Additionally, our approach outperforms probabilistic-based methods such as GeNVS \cite{chan2023genvs} and ViewSet Diffusion \cite{szymanowicz23viewset_diffusion} in FID.
Our deterministic setting generates plausible output in occluded regions, whereas ViewsetDiffusion with diffusion still yields blurry results or hallucinates implausible outcomes. For instance, the trunk of the car appears blurry and grey instead of maintaining a similar color to the front, as illustrated in Fig.~\ref{fig:main_shapenet}.

In the diffusion setting, with the incorporation of progressive inference, our model achieves even better results in distribution-based metrics (FID). However, it is worth noting that there is a slight decrease in pixel-wise metrics, namely PSNR and SSIM, due to the fact that these metrics are more favorable for mean-seeking models, as discussed in \cite{chan2023genvs}. The incorporation of a 2D diffusion model significantly reduces blurriness in the final output of the generative setting. Notably, the output of the car remains interpretable, even when considering only the input view (the last two columns in Fig.~\ref{fig:main_shapenet}). 

\begin{table}
\centering

\resizebox{\linewidth}{!}{
\begin{tabular}{lllll}
\toprule
Method      & PSNR $\uparrow$& SSIM$\uparrow$  & LPIPS$\downarrow$   & FID$\downarrow$ \\[0.1cm] \midrule 

PixelNerf                             & 23.17 & 0.90 & 0.111 &  64.08\\

VisionNerf                          & 22.87 & 0.90 & 0.084 & 24.24 \\

SSDNeRF                                & 23.52 & 0.91 & 0.078     & 16.39 \\

GeNVS                                  & 20.07 & 0.89 & 0.104  & 6.47 \\

Viewset Diffusion    & 23.29 & 0.91  & 0.094 & 39.54 \\

Splatter Image    & 24.00 & 0.92  & 0.078 & - \\
\midrule
Ours  (deterministic)       & \textbf{24.20} & \textbf{0.92}  & \textbf{0.057}  & 6.44 \\
Ours (diffusion)      & 23.67 & 0.92  & 0.061  & \textbf{6.08} \\
\bottomrule
\end{tabular}}         

\caption{Single-view novel view synthesis on ShapeNet-SRN Cars \cite{sitzmann2019srns}.}
\label{tab:main_shapenet}
\end{table}

\begin{figure}[t]
  \centering
      \centering
      \includegraphics[width=\linewidth]{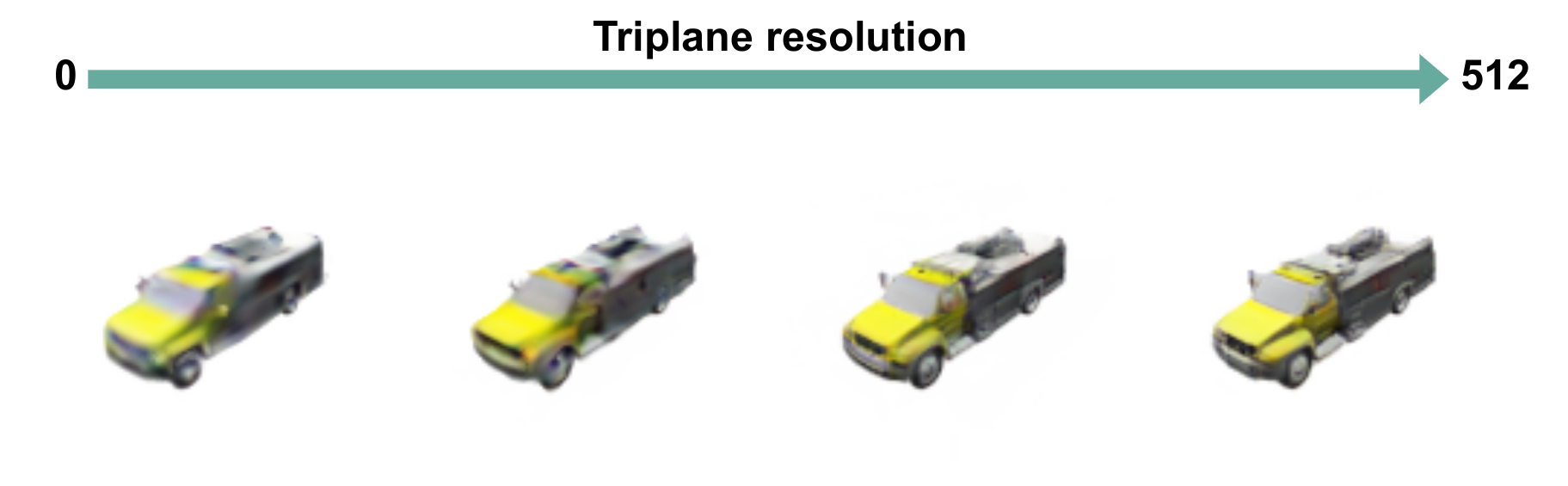}
      \caption{The effect of coarse and fine resolution of the 3D representations on final image quality (volume resolution - triplane resolution).}
      \label{fig:ablate_resolution}
  \end{figure}

\myheading{Study on Tri-plane Resolution.}
We investigate the impact of varying tri-plane resolution on reconstruction quality. We start from volumetric representation (volume dimensions equal to $16^3$) and then add a tri-plane decoder which gradually increases the tri-plane resolution. The results in Tab.~\ref{tab:ablate_resolution} demonstrate a clear trend: as the tri-plane resolution increases, the quality of novel view images improves noticeably. As shown in Fig.~\ref{fig:ablate_resolution}, the rendered images indicate enhanced sharpness as the tri-plane resolution grows higher.

\begin{table}[t]
\centering
\begin{tabular}{lccc}
    \toprule
    Triplane resolution     & PSNR $\uparrow$ & SSIM $\uparrow$  & LPIPS $\downarrow$  \\
    \midrule 
    None      & 22.78 & 0.908 & 0.087 \\
    16      & 23.07 & 0.912 & 0.081  \\
    32      & 23.53 & 0.918 & 0.069  \\
    64      & 23.70 & 0.920 & 0.078  \\
    128      & 23.76 & 0.921 & 0.067 \\
    256      & \textbf{23.79} & \textbf{0.92} & 0.066  \\
    \bottomrule
\end{tabular}
\caption{Experiments on different triplane resolution with SRN-Car dataset.}
\label{tab:ablate_resolution}
\end{table}

\section{Qualitative Results}
\label{sec:qualitative}

\myheading{Qualitative results on Shapenet SRN-Car.}
We present additional qualitative results on Shapenet SRN-Car in Table \ref{fig:supplementary_srn}. While our deterministic setting achieves good quality in the reconstruction view, it tends to produce blurry results in high ambiguity regions, like the back of the car. However, after progressive inference, our diffusion setting generates a plausible result with significantly reduced blurriness.

\begin{figure}
  \centering
  \includegraphics[width=\linewidth]{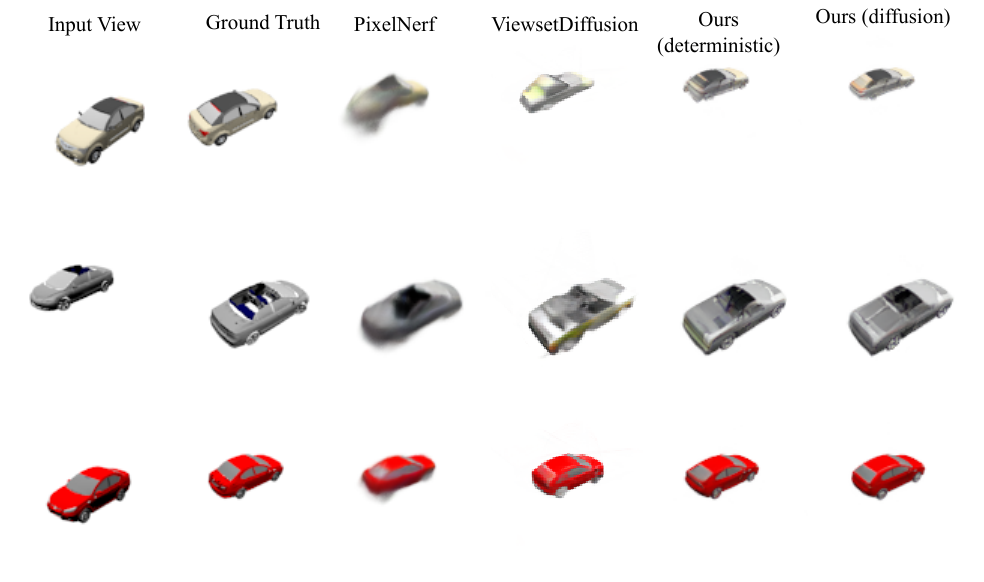}
  \caption{Qualitative results of small, non-centric sample on Shapenet SRN-Cars.}
    \label{fig:supplementary_small_srn}
\end{figure}

We include small and non-centric samples for Shapenet SRN-Car in Fig.~\ref{fig:supplementary_small_srn}. Despite ViewsetDiffusion \cite{szymanowicz23viewset_diffusion} being a probabilistic model, it still generates blurry results and saturated colors, as seen in the first row. PixelNerf \cite{yu2021pixelnerf} and our deterministic setting are regressive models, yet our method produces significantly improved results. Moreover, with progressive inference, our diffusion setting generates realistic, sharper outcomes in all samples.

\begin{figure*}[t]
  \centering
  \includegraphics[width=\linewidth]{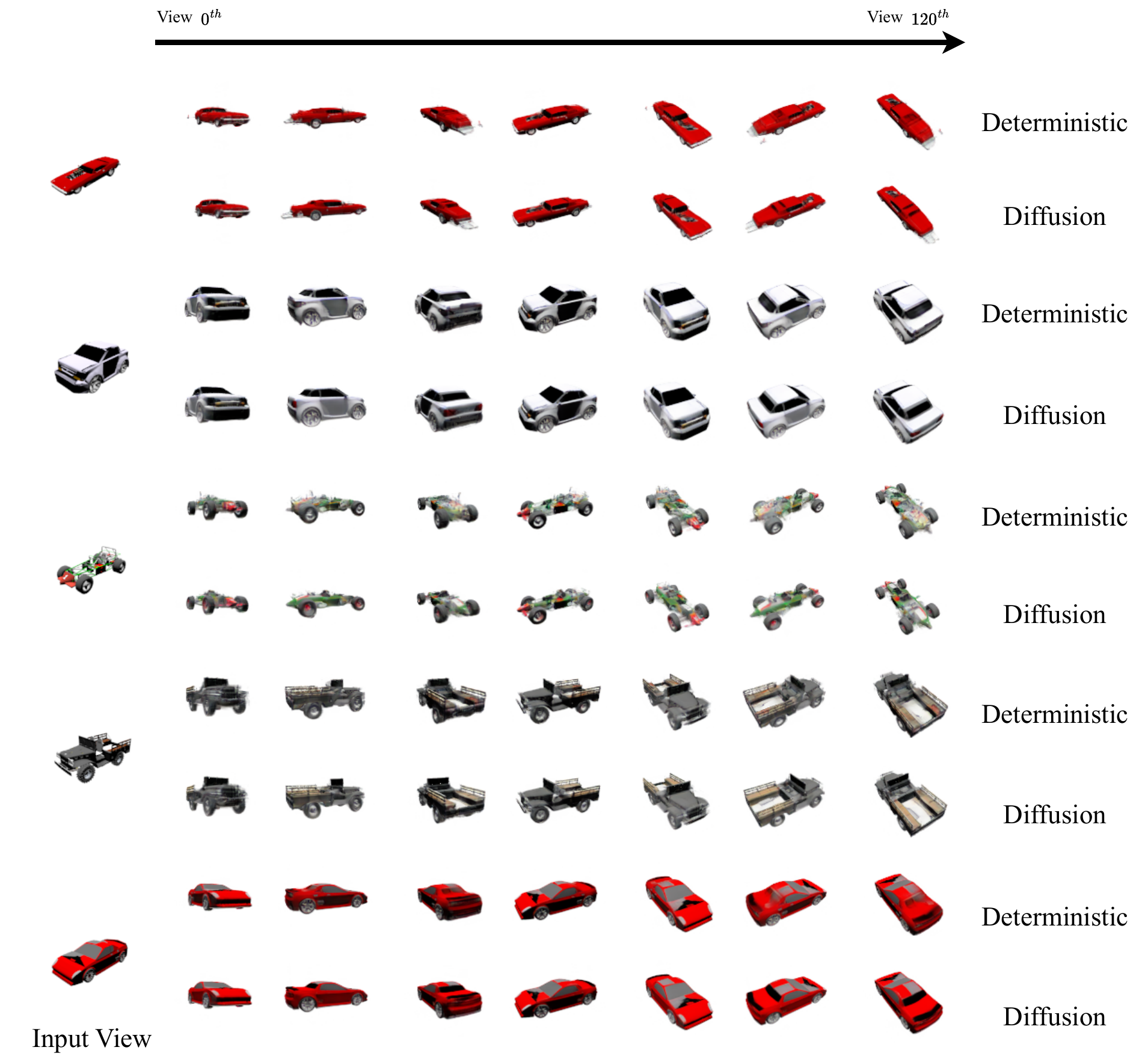}
  \caption{Qualitative results on Shapenet SRN-Cars.}
    \label{fig:supplementary_srn}
\end{figure*}

\myheading{Qualitative results on CO3D.}
We present additional qualitative results for each category of the CO3D dataset in Fig.~\ref{fig:co3d_comparison}, with Hydrant in Fig.~\ref{fig:supplementary_co3d_hydrant}, Teddybear in Fig.~\ref{fig:supplementary_co3d_teddybear}, Vase in Fig.~\ref{fig:supplementary_co3d_vase}, and Plant in Fig.~\ref{fig:supplementary_co3d_plant}. Our deterministic setting can reconstruct with mild ambiguity but tends to be blurry in high ambiguity regions, while our diffusion setting has the capability to hallucinate the unseen regions with plausible results.

While our method produces plausible results for most categories in CO3D, we found that examples in the Plant category are the most challenging, where the synthesized novel views tend to be blurry. This could be because of the high-frequency details in the plant images that are not well captured by the current resolution of our representations. Improving the high-frequency rendering of this category would be future work.

\begin{figure*}[t]
  \centering
  \includegraphics[width=\linewidth]{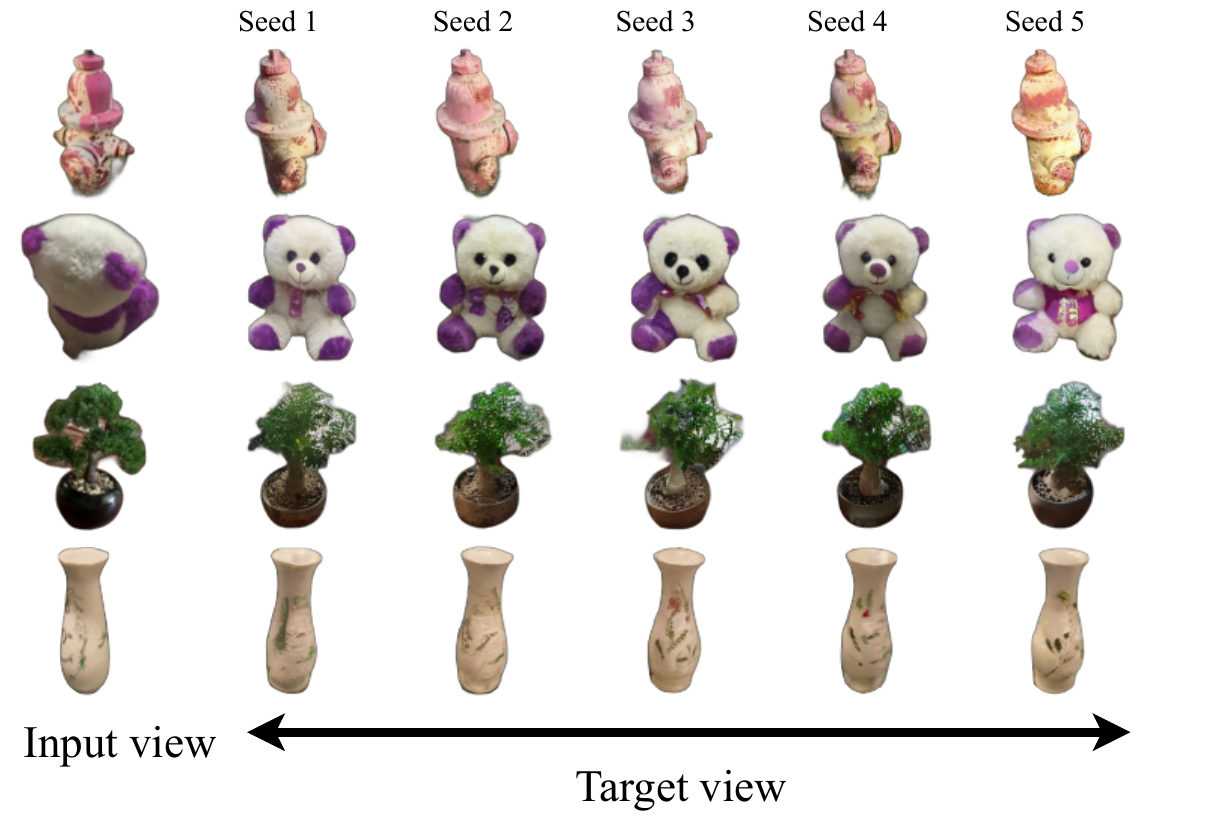}
  \caption{Sample diversity on CO3D dataset. Our diffusion model generates various feasible target views, all of which are consistent with the input view.}
    \label{fig:supplementary_co3d_variance}
\end{figure*}

\begin{figure*}[t]
  \centering
  \includegraphics[height=\pdfpagewidth]{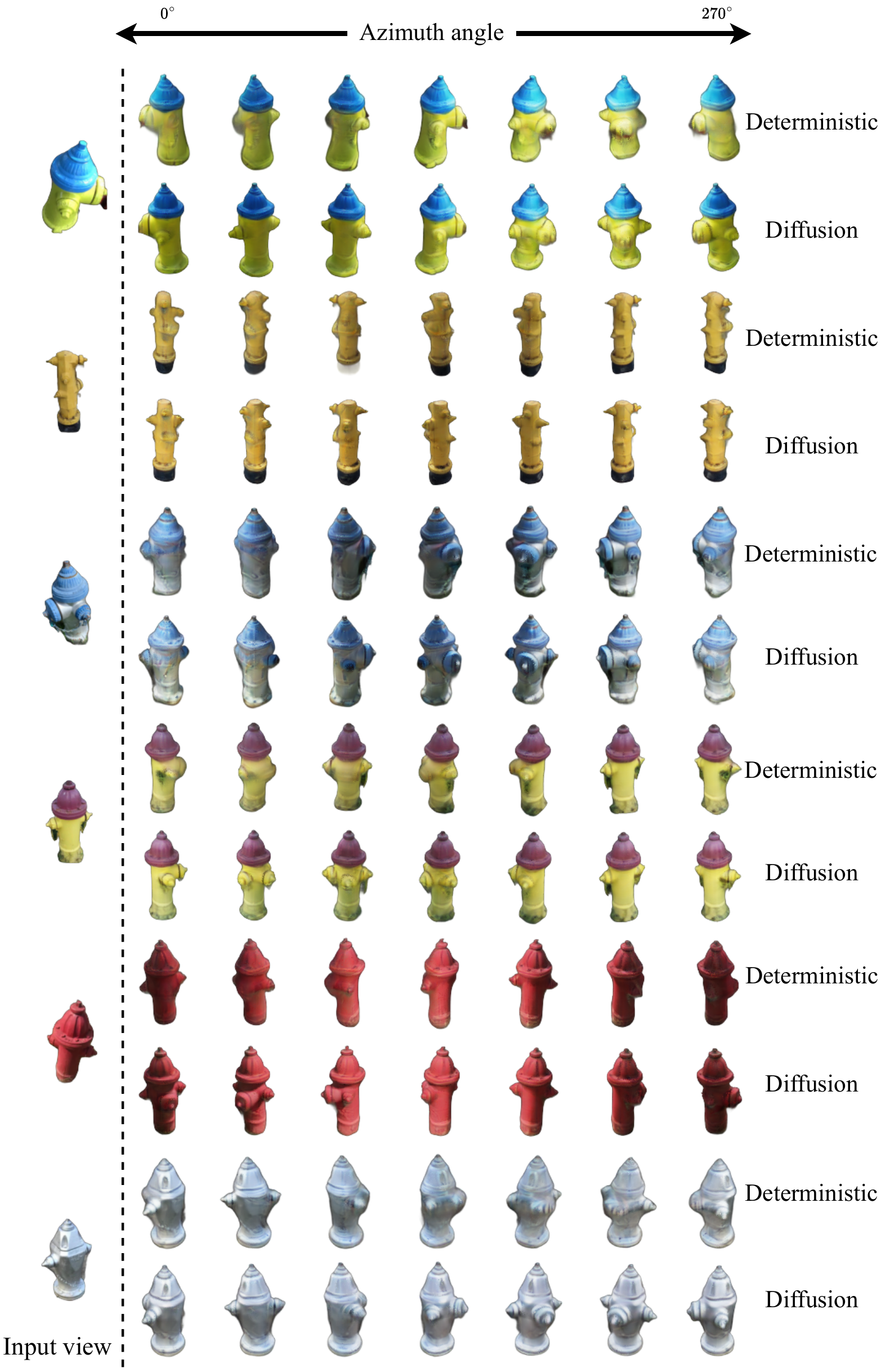}
  \caption{Qualitative results of Hydrant in CO3D dataset.}
    \label{fig:supplementary_co3d_hydrant}
\end{figure*}

\begin{figure*}[t]
  \centering
  \includegraphics[height=\pdfpagewidth]{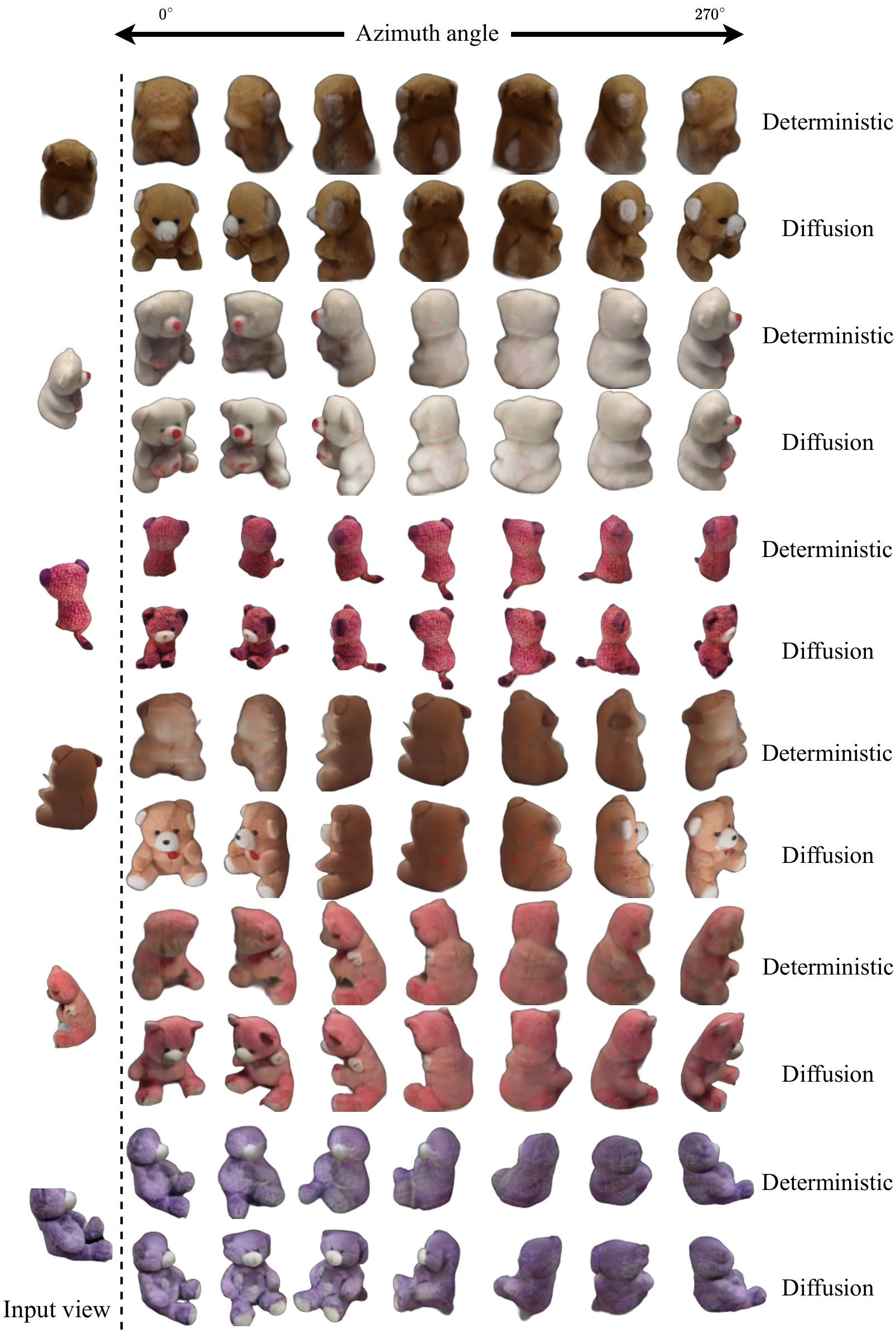}
  \caption{Qualitative results of Teddybear in CO3D dataset.}
    \label{fig:supplementary_co3d_teddybear}
\end{figure*}

\begin{figure*}[t]
  \centering
  \includegraphics[height=\pdfpagewidth]{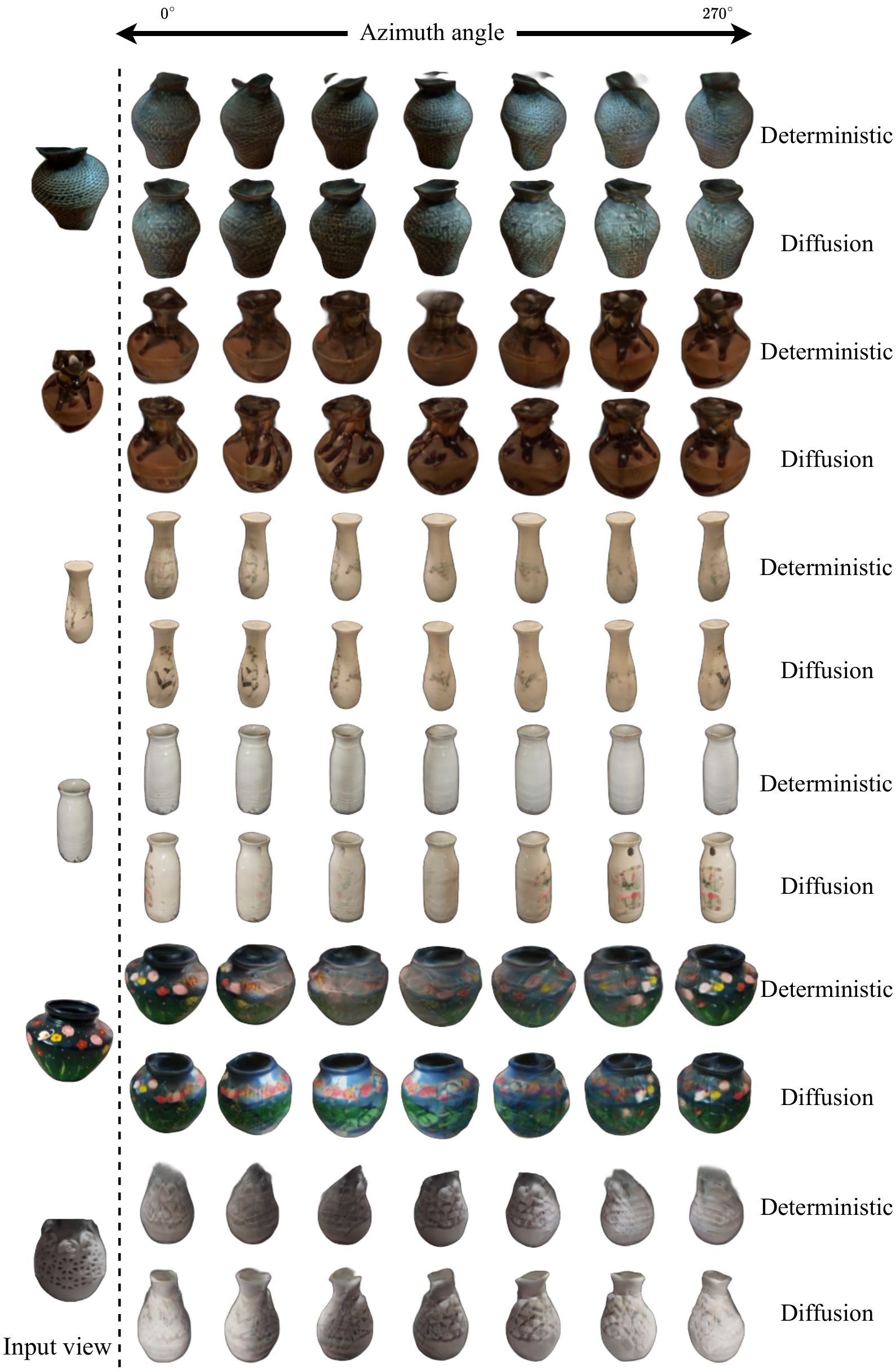}
  \caption{Qualitative results of Vase in CO3D dataset.}
    \label{fig:supplementary_co3d_vase}
\end{figure*}

\begin{figure*}[t]
  \centering
  \includegraphics[height=\pdfpagewidth]{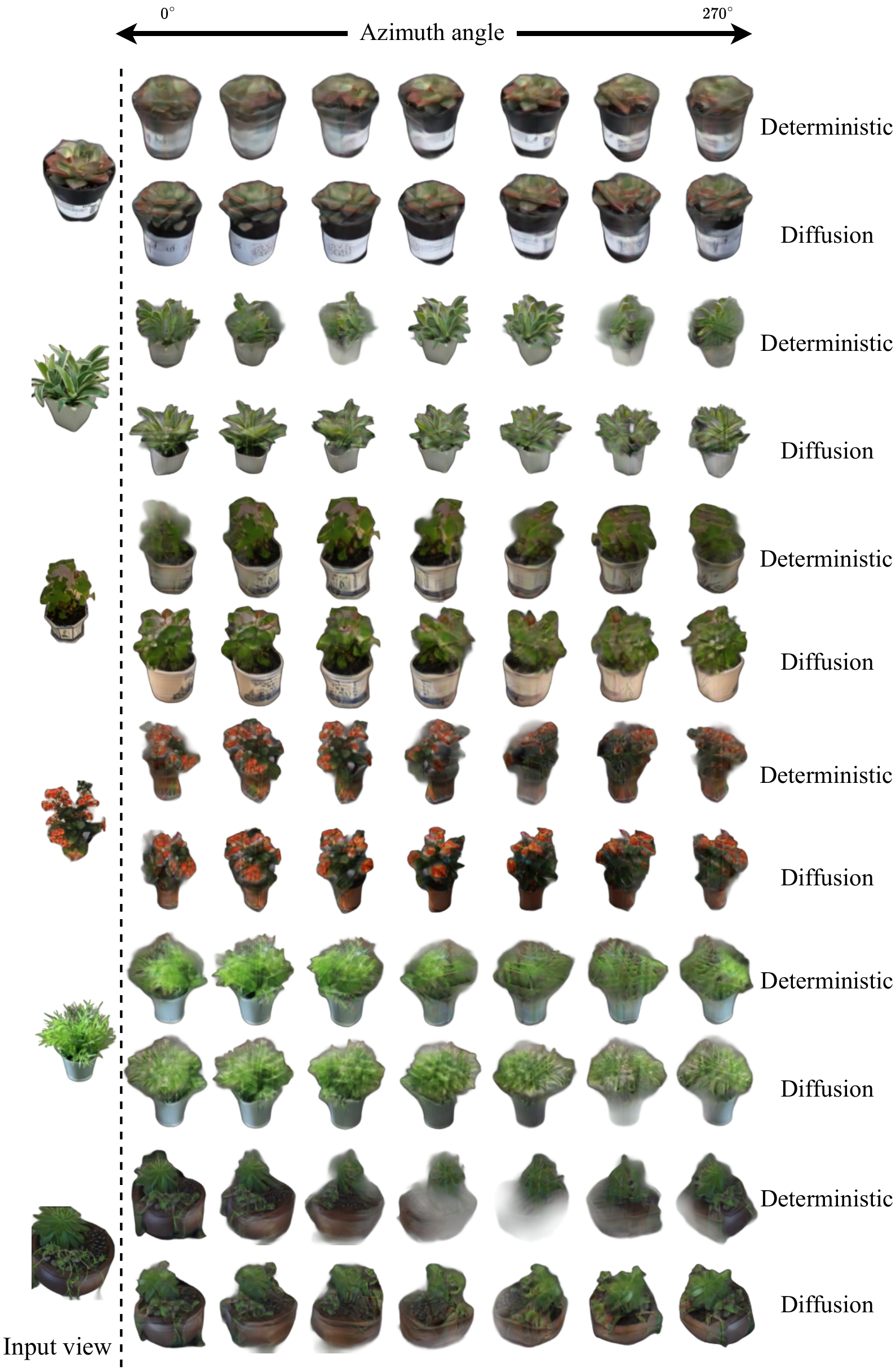}
  \caption{Qualitative results of Plant in CO3D dataset.}
    \label{fig:supplementary_co3d_plant}
\end{figure*}

\myheading{Diversity of generated images.} 
We demonstrate the diversity of the diffusion model by sampling a novel view multiple times on CO3D dataset. As shown in Fig.~\ref{fig:supplementary_co3d_variance}, each sample exhibits a distinct appearance while being consistent with the input view.

\begin{figure*}[t]
  \centering
  \includegraphics[width=\linewidth]{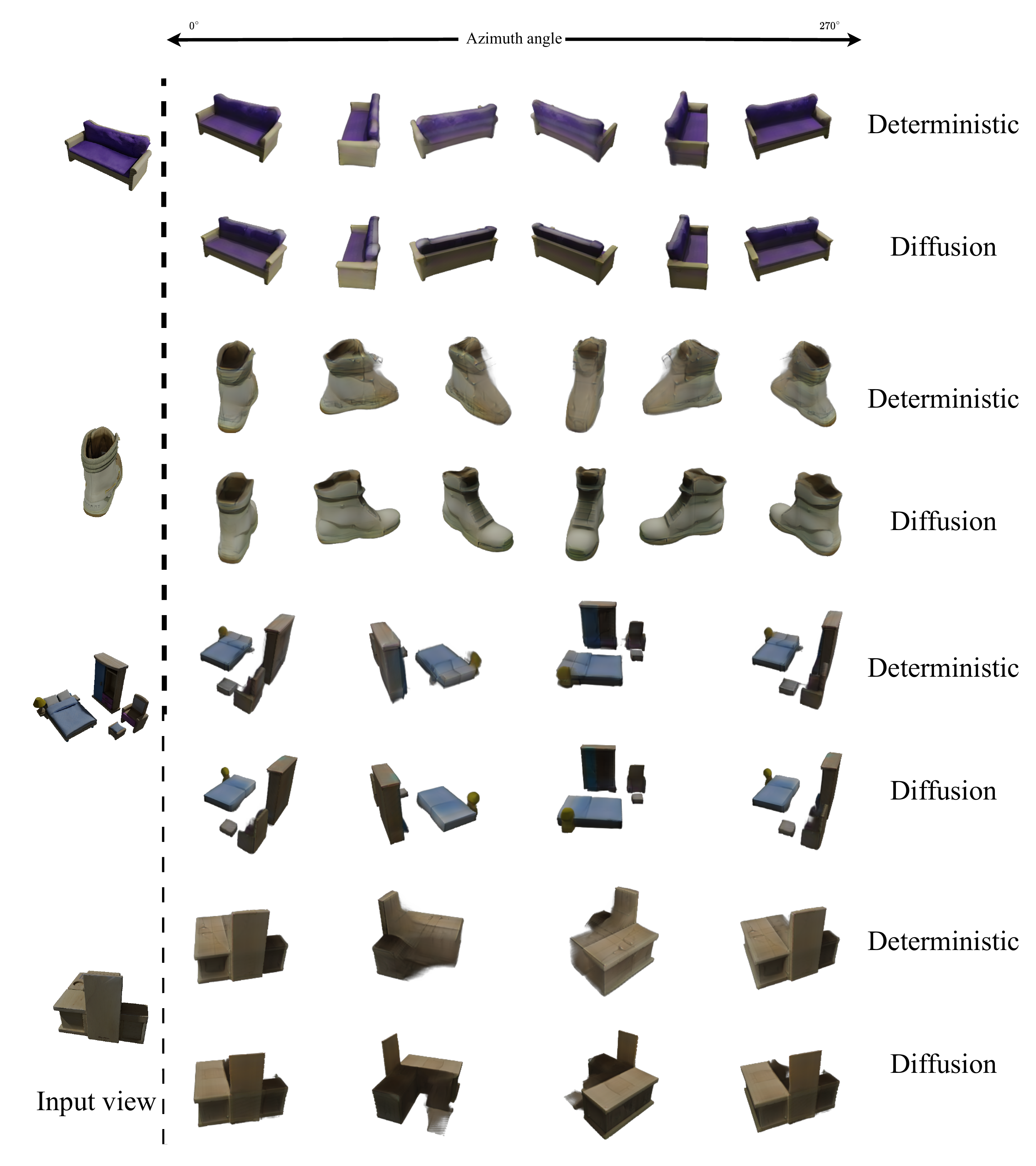}
  \caption{Single-view reconstruction results on GSO.}
    \label{fig:supplementary_gso}
\end{figure*}

\begin{figure*}[t]
  \centering
  \includegraphics[width=\linewidth]{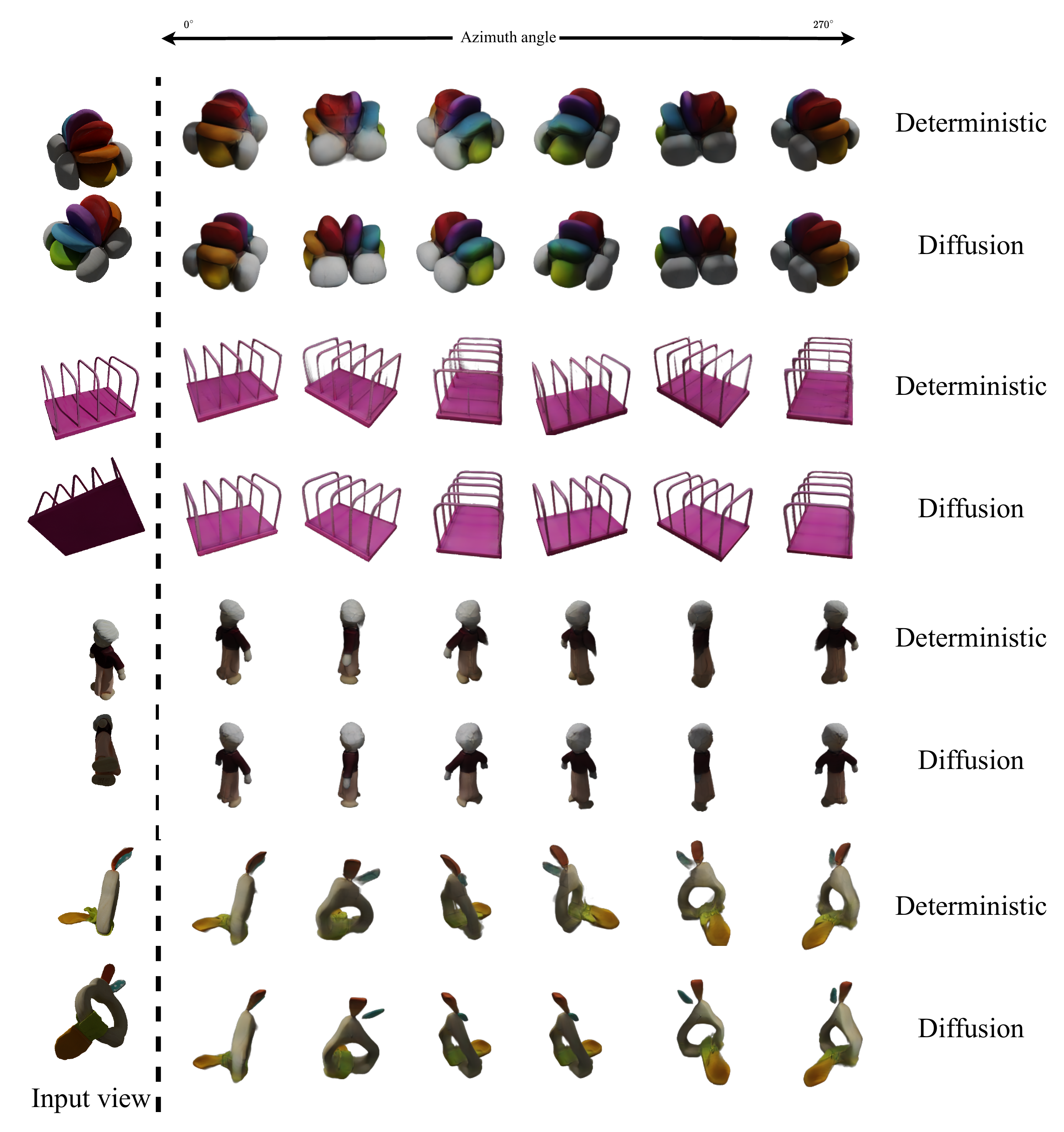}
  \caption{2-views reconstruction results on GSO.}
    \label{fig:supplementary_gso_2}
\end{figure*}

\begin{figure*}[t]
  \centering
  \includegraphics[width=\linewidth]{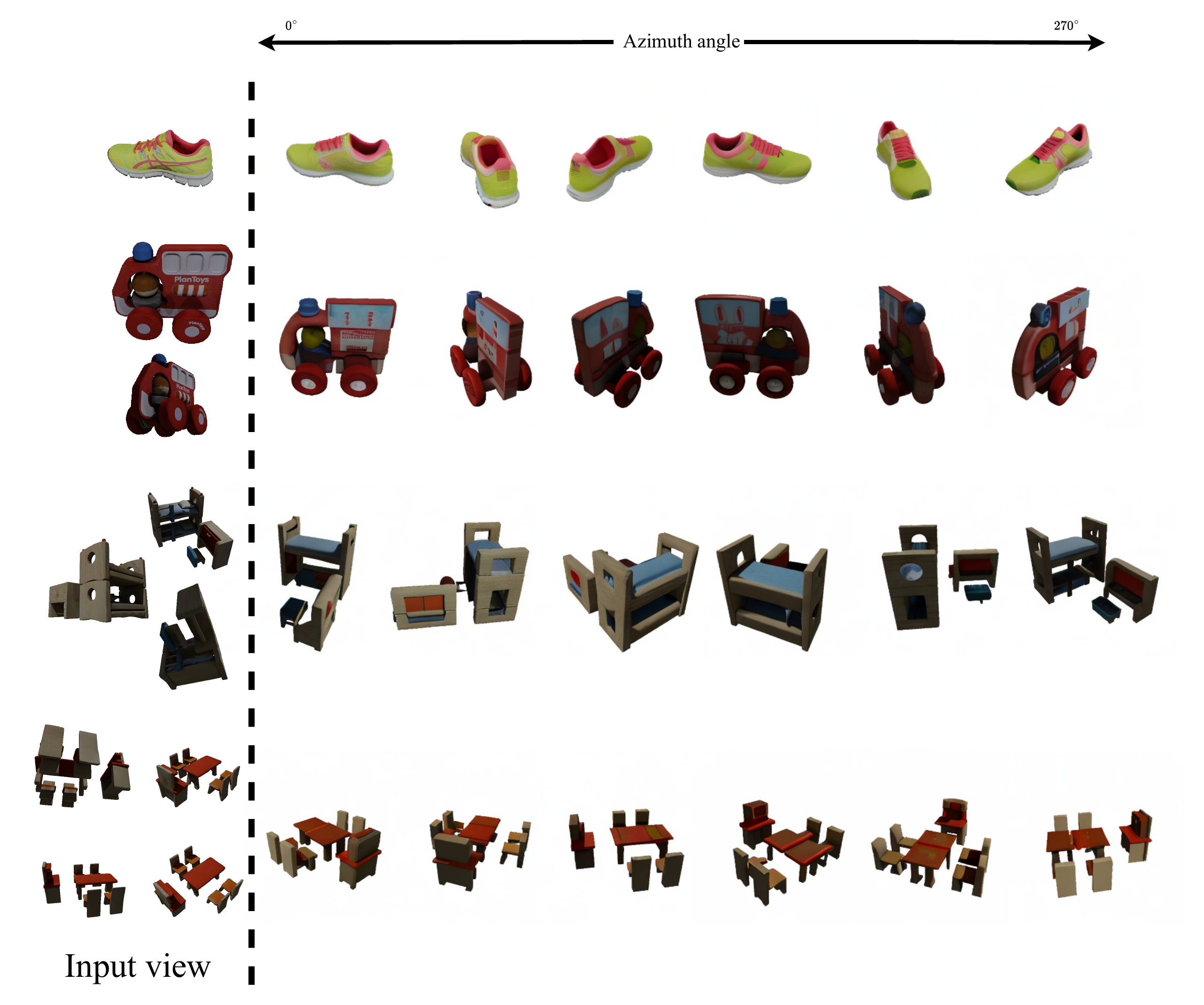}
  \caption{Sampling images from diffusion model on GSO.}
    \label{fig:supplementary_gso_intermediate}
\end{figure*}

\myheading{Qualitative results on GSO.}
We provide additional qualitative results for single-view and two-view reconstructions in Fig.\ref{fig:supplementary_gso} and Fig.\ref{fig:supplementary_gso_2}. The images generated by the diffusion model, shown in the Fig.\ref{fig:supplementary_gso_intermediate}, exhibit minor inconsistencies across views, but their fidelity is significantly higher compared to the renderings produced by our 3D representation.

\myheading{Comparision with SyncDreamer on GSO dataset.}
We compare our deterministic model with the multi-view diffusion method, SyncDreamer \cite{liu2023syncdreamer}, in Table \ref{tab:main_syncdreamer}. While our approach demonstrates competitive performance against SyncDreamer, it also offers greater flexibility. Unlike SyncDreamer, which generates only 16 views at a fixed elevation of 0 degrees, our method can accept arbitrary input view(s) and render at any desired pose(s).

\begin{table}[H]
\centering

\resizebox{0.8\linewidth}{!}{
\begin{tabular}{lllll}
\toprule
Method      & PSNR $\uparrow$& SSIM$\uparrow$  & LPIPS$\downarrow$  \\ 
\midrule 

SyncDreamer       & 19.46 & \textbf{0.83}  & \textbf{0.142}  \\
Ours      & \textbf{19.49} & 0.82  & 0.180 \\
\bottomrule
\end{tabular}}         

\caption{Comparison with SyncDreamer on GSO dataset. While SyncDreamer produces perceptually better results, it can only generates a fixed set of 16 views at fixed camera angles from a frontal input view. By contrast, our method can generate novel views with input/output at arbitrary camera angles.}
\label{tab:main_syncdreamer}
\end{table}

\clearpage

\end{document}